\documentclass[accepted]{uai2025} 
                        
\usepackage[american]{babel}

\usepackage{natbib} 
\usepackage{mathtools} 
\usepackage{booktabs} 
\usepackage{tikz} 

\usepackage{amssymb}
\usepackage{algorithm}
\usepackage{algpseudocode}
\usepackage{multirow}
\usepackage[FIGTOPCAP]{subfigure}

\newtheorem{proposition}{Proposition}
\newtheorem{lemma}{Lemma}
\newtheorem{theorem}{Theorem}

\title{Fast Calculation of Feature Contributions in Boosting Trees}

\author[1]{{Zhongli Jiang}{}}
\author[1]{{Min Zhang}{}}
\author[1]{{Dabao Zhang}{}}

\affil[1]{%
    Department of Epidemiology \& Biostatistics\\
    University of California\\
    Irvine, CA, USA
}

\begin{document}
\maketitle

\begin{abstract}
Recently, several fast algorithms have been proposed to decompose predicted value into Shapley values, enabling individualized feature contribution analysis in tree models. While such local decomposition offers valuable insights, it underscores the need for a global evaluation of feature contributions. Although coefficients of determination ($R^2$) allow for comparative assessment of individual features, individualizing $R^2$ is challenged by the underlying quadratic losses. To address this, we propose Q-SHAP, an efficient algorithm that reduces the computational complexity of calculating Shapley values for quadratic losses to polynomial time. Our simulations show that Q-SHAP not only improves computational efficiency but also enhances the accuracy of feature-specific $R^2$ estimates.
\end{abstract}

\section{Introduction} \label{introduction}
\label{sec:intro}

Models built with tree ensembles are powerful but often complicated, making it challenging to understand the influence of inputs. Feature importance plays a critical role in demystifying these models and enhancing their interpretability by assigning each input feature a score. This is crucial in domains like healthcare and biomedicine, where trust and interpretation of the model are essential \citep{stiglic2020interpretability, bussmann2021explainable}. Common feature importance measures like gain can be inconsistent \citep{lundberg2018consistent} while permutation importance lacks theoretical foundations \citep{ishwaran2007variable}.  

Shapley values, derived from cooperative game theory and introduced by \citet{shapley1953value}, offer a robust method for the fair distribution of payoffs generated by a coalition of players. This can be analogously applied to assess the contribution of each feature in a machine learning model. It ensures that each feature's contribution is assessed by considering all possible combinations of features, thereby providing a comprehensive understanding of feature impacts. Recent applications of Shapley values have focused on local interpretation \citep{lundberg2017unified, lundberg2018consistent, chau2022rkhs}, where they are employed to examine the influence of individual features on specific predictions. Nonetheless, there are numerous scenarios where global importance is preferred, such as analysis of the role of a feature across the entire dataset \citep{molnar2020interpretable, covert2020understanding}.

Among the works that compute Shapley values in a global context, a popular approach is to use model variance decomposition. \citet{lipovetsky2001analysis} decomposed $R^2$ in linear regression, offering consistent interpretations even in the presence of multicollinearity. \citet{owen2017shapley} also conducted a conceptual analysis of Shapley values for the variance. However, computation remains a significant challenge, as the calculation of Shapley values grows exponentially with the number of features. To address this issue, several Monte Carlo-based methods have been proposed to effectively reduce the computational burden \citep{song2016shapley, covert2020understanding, williamson2020efficient}.

Although Monte Carlo-based, model-agnostic methods are more efficient than brute-force approaches, they are still computationally intensive, especially when dealing with high-dimensional data that requires extensive feature permutation sampling to ensure consistency \citep{lundberg2017unified, lundberg2020local}. This challenge has prompted the development of methods that leverage the specific structures of tree-based models. However, much of the focus has been on explaining individual predictions, as seen with TreeSHAP \citep{lundberg2018consistent}, FastTreeSHAP \citep{yang2021fast}, LinearTreeSHAP \citep{bifet2022linear} and FourierSHAP \citep{gorji2024amortized}. \citet{benard2022shaff} considered population-level importance using $R^2$, specifically tailored for random forests \citep{breiman2001random}.

\citet{lundberg2020local} suggests that explaining the loss function for a ``path-dependent'' algorithm is challenging. To the best of our knowledge, there is no available method to calculate Shapley values of quadratic losses by leveraging structures of decision trees for fast computation. In this paper, we propose Q-SHAP, which can decompose quadratic terms of predicted values of a decision tree into each feature's attribute in polynomial time. It leads to fast computation of feature-specific $R^2$ for a decision tree. We also extend our approach to Gradient Boosted Decision Trees.

The rest of the paper is structured as follows. In Section~\ref{intro}, we provide a brief overview of Shapley values of $R^2$. In Section~\ref{qshap_algo}, we present our proposed algorithm Q-SHAP to calculate Shapley values of $R^2$ in polynomial time for single trees, and then extend the approach for tree ensembles in Section~\ref{qshap-booting}. We justify the efficacy and efficiency of the algorithm using extensive simulations in Section~\ref{simulation_study} and real data analysis in high dimension in Section~\ref{real}. We conclude the paper with a discussion in Section~\ref{conclusion}.



\section{Shapley Values of \texorpdfstring{$R^2$}{R-squared} for Individual Features} \label{intro}

\subsection{Model specification}

Here we investigate a specific label $Y$ and its explainability by a full set of $p$ features $X=(X_1, X_2, \cdots, X_p)$. For any subset $F\subseteq \mathcal{P}=\{1, 2, \cdots, p\}$, we define the corresponding set of features as $X_F=(X_j)_{j\in F}$.

Suppose that, for any set of features $X_F$, an oracle model $m_F$ can be built such that, for any value $x=(x_j)_{j\in\mathcal{P}}$, 
\[
m_F(x) = E[Y|X_F=(x_j)_{j\in F}].
\]

The Shapley value of $j$-th feature, in terms of its contribution to the total variation, is defined as
\begin{eqnarray} \label{rsq_definition}
\phi_{\rho^2,j} &=& \frac{1}{p~var(m_{\emptyset})} \sum_{F\subseteq\mathcal{P}\backslash\{j\}} 
\begin{pmatrix} p-1\\|F|\end{pmatrix}^{-1} 
\nonumber\\ && \times
\left(var(m_{F\cup\{j\}}) - var(m_F)\right),
\end{eqnarray}
where $|F|$ is the number of features in $F$. The term $var(m_{F\cup\{j\}})$ is the variance explained by feature set $F\cup\{j\}$ and the term $var(m_F)$ is the variance explained solely by set $F$.
This definition is analogous to \citet{covert2020understanding} and \citet{williamson2020efficient}. By averaging over all possible feature combinations, the Shapley values are the only solution that satisfies the desired properties of symmetry, efficiency, additivity, and dummy \citep{shapley1953value}.

\subsection{Empirical Estimation}

Suppose we have a set of data with sample size $n$ observed for both label and features as
\begin{eqnarray*}
\mathbf{Y} &=& (y_1, y_2, \cdots, y_n),\\
\mathbf{X} &=& \left(\mathbf{X}_{\cdot 1}, \mathbf{X}_{\cdot 2}, \cdots, \mathbf{X}_{\cdot p}\right) = \left(\mathbf{x}_{1\cdot}^T, \mathbf{x}_{2\cdot}^T, \cdots, \mathbf{x}_{n\cdot}^T\right)^T.
\end{eqnarray*}
Accordingly, we denote the observed data of features in subset $F$ as 
\[
\mathbf{X}_{\cdot F}=(\mathbf{X}_{\cdot j})_{j\in F}.
\]

Suppose that, for each subset $F$ of features, a single optimal model $\hat{m}_F$ is built on data $(\mathbf{Y}, \mathbf{X}_{\cdot F})$. Then the $i$-th label can be predicted with 
\[
\hat{y}_i(\mathbf{X}_{\cdot F}) = \hat{m}_F(\mathbf{x}_{i\cdot}). 
\]

\subsection{From \texorpdfstring{$R^2$}{R-squared} to a Quadratic Loss}\label{rsq2dsq}

We will establish the connection of $R^2$ to a quadratic loss through equation (\ref{rsq_definition}). We define the quadratic loss on the optimal model $\hat{m}_F$ as 
\begin{eqnarray}\label{eqn_qloss}
Q_F=\sum_{i=1}^{n}\left(y_i-\hat{m}_F(\mathbf{x}_{i\cdot})\right)^2
\end{eqnarray}
for any set of features $F$. With $m_{\emptyset}(\mathbf{x}_{i\cdot})=\bar{y}$, we have $Q_{\emptyset}=\sum_{i=1}^{n}(y_i-\bar{y})^2.$ Following the law of total variance, we can estimate $var(m_F)$ by
\[
\widehat{var}(m_F) = \left(Q_{\emptyset}-Q_F\right)\big/n.
\]
Thus, an empirical estimate of (\ref{rsq_definition}) is
\begin{eqnarray*}
\phi_{R^2,j} 
=-\frac{1}{p Q_{\emptyset}}
\sum_{F\subseteq\mathcal{P}\backslash\{j\}} \begin{pmatrix} p-1\\ |F|\end{pmatrix}^{-1} 
(Q_{F\cup\{j\}}-Q_{F}), 
\end{eqnarray*}
which is proportional to a Shapley value for the sum of squared errors, i.e., the quadratic loss in (\ref{eqn_qloss}).

\subsection{From Quadratic Loss to Q-SHAP}

We now further reduce Shapley values of the sum of squared errors to Shapley values of linear and quadratic terms of predicted values. Expanding the loss function in (\ref{eqn_qloss}), we can rewrite, 
\begin{eqnarray}
\phi_{R^2,j}  
&=& -\frac{1}{p Q_{\emptyset}}\sum_{F\subseteq\mathcal{P}\backslash\{j\}}\begin{pmatrix} p-1\\ |F|\end{pmatrix}^{-1} \sum_{i=1}^{n} \left(\hat{m}_{F\cup j}^2(\mathbf{x}_{i\cdot})\right.\nonumber \\
&& \left.-\hat{m}_F^2(\mathbf{x}_{i\cdot}) - 2(\hat{m}_{F\cup j}(\mathbf{x}_{i\cdot}) - \hat{m}_F(\mathbf{x}_{i\cdot}))y_i\right). \nonumber 
\end{eqnarray}

To calculate this, we define the Shapley value for each sample $i$ as,
\setlength{\arraycolsep}{2pt}
\begin{eqnarray*}
\lefteqn{\phi_{R^2,j}(\mathbf{x}_{i\cdot})}\\ 
&=&-\frac{1}{p Q_{\emptyset}} \sum_{F\subseteq\mathcal{P}\backslash\{j\}} \begin{pmatrix} p-1\\ |F|\end{pmatrix}^{-1} \left(\hat{m}_{F\cup j}^2 (\mathbf{x}_{i\cdot}) - \hat{m}_F^2(\mathbf{x}_{i\cdot})\right) \\
&&+ \frac{2y_i}{p Q_{\emptyset}} \sum_{F\subseteq\mathcal{P}\backslash\{j\}} \begin{pmatrix} p-1\\ |F|\end{pmatrix}^{-1} \left(\hat{m}_{F\cup j}(\mathbf{x}_{i\cdot}) - \hat{m}_F(\mathbf{x}_{i\cdot}) \right),
\end{eqnarray*}

which is a linear combination of two sets of Shapley values, i.e., Shapley values of predicted value $\hat{m}_F$, which are ready to be calculated \citep{lundberg2018consistent, yang2021fast, bifet2022linear}, and Shapley values of the quadratic term of predicted value $\hat{m}_F^2$, i.e.,
\begin{eqnarray}\label{msq}
\phi_{\hat{m}^2,j}(\mathbf{x}_{i\cdot})  &=&
\frac{1}{p} \sum_{F\subseteq\mathcal{P}\backslash\{j\}} \begin{pmatrix} p-1\\ |F|\end{pmatrix}^{-1} \nonumber\\
&& \times \left(\hat{m}_{F\cup j}^2(\mathbf{x}_{i\cdot}) 
 - \hat{m}_F^2(\mathbf{x}_{i\cdot})\right),
\end{eqnarray}
for which we will develop the algorithm Q-SHAP to calculate. 
For the rest of the paper, we will focus on computing the Shapley values in Equation~{(\ref{msq})} in polynomial time for tree-based models and carrying it over to calculate feature-specific $R^2$.

\section{The Algorithm Q-SHAP for Single Trees} \label{qshap_algo}

Here we will focus on calculating the Shapley values in~(\ref{msq}) for a single tree. By definition, each Shapley value~(\ref{msq}) requires evaluating all feature subsets, leading to an NP-hard computation in general. However, binary trees restrict predictions to a finite set of values attached to leaf nodes, and we will show that summarizing differences of feature subsets are related to operations on polynomial coefficients. Leveraging our results of polynomial identity in Proposition~\ref{dimension_reduce} and Proposition~\ref{product}, we can calculate this Shapley value over pairs of leave nodes, shown in Theorem~\ref{T2}, instead of all feature subsets.

\subsection{Notations}

We assume the underlying decision tree has the maximum depth at $D$ and a total of $L$ leaves, and use $l$ to denote a specific leaf. We further introduce a dot product for polynomials for subsequent calculation. For two polynomials $A(z) = \sum_{i=0}^{n} a_i z^i$ and $B(z) = \sum_{i=0}^{n} b_i z^i$, we define their dot product as $A(z) \cdot B(z) = \sum_{i=0}^{n} a_i b_i$.

\subsection{Distributing the prediction to leaves}

Decision trees match each data point to one leaf for prediction. However, for our prediction defined on any subset $F$, a data point $\mathbf{x}_{i\cdot}$ can fall into multiple leaves due to the uncertainty by unspecified features $\mathcal{P}\backslash F$. We can calculate $\hat{m}_F(\mathbf{x}_{i\cdot})$, following TreeSHAP, as the empirical mean by aggregating the weighted prediction on each leaf,
\begin{equation}\label{linear_leaf}
\hat{m}_F(\mathbf{x}_{i\cdot}) = \sum_l 
\hat{m}_F^l(\mathbf{x}_{i\cdot}) \triangleq \sum_l 
w(\mathbf{x}_{i\cdot}, l, F) \times \hat{m}^l,
\end{equation}
where $\hat{m}^l$ is the predicted value at leaf $l$ based on the model built on all features.

\begin{figure}[ht]
    \centering
    \begin{minipage}[b]{0.48\textwidth}
        \centering (a) Decision tree built on $X_1$ and $X_2$
    \end{minipage}
    \begin{minipage}[b]{0.48\textwidth}
        \centering
        \includegraphics[scale=0.7]{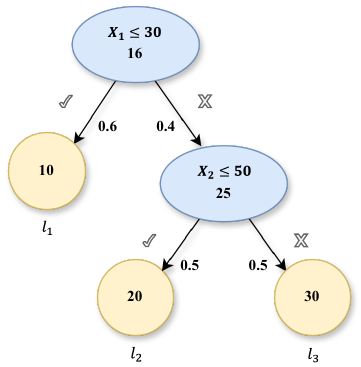}
    \end{minipage}
    \hfill

    \begin{minipage}[b]{0.48\textwidth}
    \centering (b) Hypothetical tree with $X_1$ only
    \end{minipage}
    \begin{minipage}[b]{0.48\textwidth}
        \centering
        \includegraphics[scale=0.7]{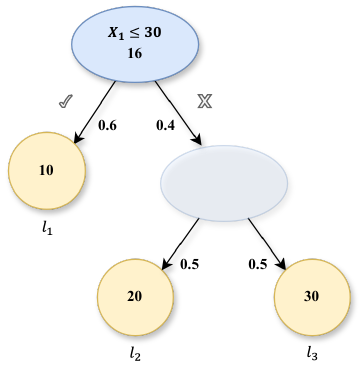}
    \end{minipage}

     \caption{Illustration of (a) a decision tree built with both features $X_1$ and $X_2$ and (b) its hypothetical tree with only feature $X_1$.} \label{illu_tree}
\end{figure}

Given an oracle tree built on all available features, we try to recover the oracle tree for a subset of features without rebuilding, following \citet{bifet2022linear} and \citet{karczmarz2022improved}. For example, suppose the tree involves two features, $X_1$ and $X_2$, as shown in Figure~\ref{illu_tree}.a. However, when feature $X_2$ is excluded from the structure, we would use only $X_1$ to build the tree as shown in Figure~\ref{illu_tree}.b.

Therefore, we replace it with a pseudo internal node to preserve the structure of the original full oracle tree and pave the way for further formulation.

With a data point $\mathbf{x}_{i\cdot}=(40,25)$, we illustrate the calculation in Equation~{(\ref{linear_leaf})} by first calculating the predicted value for the tree in Figure~\ref{illu_tree}.b, $\hat{m}_{\{1\}}(\mathbf{x}_{i\cdot})= \hat{m}_{\{1\}}^{l_1}(\mathbf{x}_{i\cdot})+\hat{m}_{\{1\}}^{l_2}(\mathbf{x}_{i\cdot})+\hat{m}_{\{1\}}^{l_3}(\mathbf{x}_{i\cdot}) = 0 \times 10 + 0.5 \times 20 + 0.5 \times 30$.

When the tree is built with an additional feature $j=2$ as shown in Figure~\ref{illu_tree}.a, we have the predicted value $\hat{m}_{\{1, 2\}}(\mathbf{x}_{i\cdot})= \hat{m}_{\{1, 2\}}^{l_1}(\mathbf{x}_{i\cdot}) + \hat{m}_{\{1, 2\}}^{l_2}(\mathbf{x}_{i\cdot}) + \hat{m}_{\{1, 2\}}^{l_3}(\mathbf{x}_{i\cdot}) =\textbf{1} \times 0 \times 10 + \mathbf{0.5^{-1}} \times 0.5 \times 20 + \textbf{0} \times 0.5 \times 30$

where the bold numbers reweight $\hat{m}^l_{\{1\}}(\mathbf{x}_{i\cdot})$ for $\hat{m}^l_{\{1, 2\}}(\mathbf{x}_{i\cdot})$. Let us take a closer look at these weights, each corresponding to one leaf. For leaf $l_1$, the weight is 1 since the newly added feature $X_2$ is not involved in its path and the reweighted prediction remains as zero. For leaf $l_2$, the reweighted prediction is lifted up with the weight inversely proportional to the previous probability because $\mathbf{x}_{i\cdot}$ follows its path to the leaf with probability 1. On the other hand, although the path to leaf $l_3$ includes the newly added feature, $\mathbf{x}_{i\cdot}$ doesn't follow this path, resulting in a weight at 0. Next, we will generalize such a reweighting strategy to calculate (\ref{linear_leaf}) for trees with different sets of features. 

Denote $F^l$ the features involved in the path to leaf $l$ and $F^l(\mathbf{x}_{i\cdot})$ the subset of $F^l$ whose decision criteria are satisfied by $\mathbf{x}_{i\cdot}$. Note that each feature $j\in F^l$ may appear multiple times in the path to leaf $l$ so we denote $n^l_{j, c}$ the number of samples passing through the node which is attached to the $c$-th appearance. We similarly define $n^l_{j, c}(\mathbf{x}_{i\cdot})$ for each feature $j\in F^l(\mathbf{x}_{i\cdot})$.

For any feature $j \in \mathcal{P}$, we can define the weight function based on a partition of $\mathcal{P}$ into three subsets $F^l(\mathbf{x}_{i\cdot})$, $F^l \backslash F^l(\mathbf{x}_{i\cdot})$, and $\mathcal{P}\backslash F^l$,
\begin{equation*} 
 w_j^l(\mathbf{x}_{i\cdot}) \triangleq 
\begin{cases} 
\prod_c{\frac{n^l_{j, c}}{n^l_{j,c}(\mathbf{x}_{i\cdot})}}, & \text{if } j \in F^l(\mathbf{x}_{i\cdot}); \\
0, & \text{if } j \in F^l \backslash F^l(\mathbf{x}_{i\cdot}); \\
1, & \text{if } j \in \mathcal{P}\backslash F^l.
\end{cases}
\end{equation*}
Therefore, for $j \notin F$, we have
\begin{equation*}
\hat{m}_{F\cup j}^l(\mathbf{x}_{i\cdot})=w_j^l(\mathbf{x}_{i\cdot})\hat{m}_{F}^l(\mathbf{x}_{i\cdot}).
\end{equation*}
Recursive application of the above formula leads to 
\begin{equation*}
\hat{m}_{F}^l(\mathbf{x}_{i\cdot})=\prod_{k\in F}w^l_k(\mathbf{x}_{i\cdot})\hat{m}_{\emptyset}^l,
\end{equation*}
where $\hat{m}_{\emptyset}^l=\hat{m}^l\frac{n^l}{n}$ with $n^{l}$ the sample size at leaf $l$, $n$ the total sample size.

When $F=\emptyset$, the above result reduces to $\hat{m}_{\emptyset}(\mathbf{x}_{i\cdot}) = \sum_l\hat{m}^l\frac{n^l}{n},$ so the optimal prediction is just the mean for all data points, which is consistent with $\hat{m}_{\emptyset}(\mathbf{x}_{i\cdot})=\bar{y}$.

We can rewrite (\ref{msq}) by aggregating over leaves as
\begin{eqnarray}\label{mseqsplit}
\lefteqn{\phi_{\hat{m}^2,j}(\mathbf{x}_{i\cdot})} \nonumber\\
&=& \frac{1}{p} \sum_{F\subseteq\mathcal{P}\backslash\{j\}} \begin{pmatrix} p-1\\ |F|\end{pmatrix}^{-1} \nonumber \\
&& \times\left(\sum_l \left(w_j^{l2}(\mathbf{x}_{i\cdot})-1\right) \hat{m}^{l2}_{\emptyset}\prod_{k\in F}w_k^{l2}(\mathbf{x}_{i\cdot})\right) \nonumber \\
&& + \frac{2}{p} \sum_{F\subseteq\mathcal{P}\backslash\{j\}} \begin{pmatrix} p-1\\ |F|\end{pmatrix}^{-1} \nonumber \\
&& \times \left( \sum_{l_1\neq l_2}(w_j^{l_1}(\mathbf{x}_{i\cdot})w_j^{l_2}\left(\mathbf{x}_{i\cdot})-1\right)\hat{m}_{\emptyset}^{l_1}\hat{m}^{l_2}_{\emptyset}\right.\nonumber\\
&& \times\left.\prod_{k\in F}w_k^{l_1}(\mathbf{x}_{i\cdot})w_k^{l_2}(\mathbf{x}_{i\cdot})\right) \nonumber\\
&\triangleq& T_{1,j}(\mathbf{x}_{i\cdot}) + 2 T_{2,j}(\mathbf{x}_{i\cdot}).
\end{eqnarray}

We further define, for leaves $l_1$ and $l_2$,
\begin{eqnarray}\label{eqn_T2}
\lefteqn{T_{j}^{l_1l_2}(\mathbf{x}_{i\cdot})} \nonumber\\
&=& \frac{1}{p} \sum_{F\subseteq\mathcal{P}\backslash\{j\}} \begin{pmatrix} p-1\\ |F|\end{pmatrix}^{-1} \bigg( (w_j^{l_1}(\mathbf{x}_{i\cdot})w_j^{l_2}(\mathbf{x}_{i\cdot})-1) \nonumber\\
&& \times \hat{m}_{\emptyset}^{l_1}\hat{m}^{l_2}_{\emptyset} \prod_{k\in F}w_k^{l_1}(\mathbf{x}_{i\cdot})w_k^{l_2}(\mathbf{x}_{i\cdot})\bigg),
\end{eqnarray}
and we have $T_{2,j}(\mathbf{x}_{i\cdot})=\sum_{l_1 \neq l_2}T_{j}^{l_1l_2}(\mathbf{x}_{i\cdot})$, $T_{1,j}(\mathbf{x}_{i\cdot})=\sum_{l}T_{j}^{ll}(\mathbf{x}_{i\cdot}).$
Therefore, we will focus on the calculation of $T_{j}^{l_1l_2}(\mathbf{x}_{i\cdot})$ in (\ref{eqn_T2}) the rest of this section.

We can reduce the calculation of $\prod_{k\in F}w_k^{l_1}(\mathbf{x}_{i\cdot})w_k^{l_2}(\mathbf{x}_{i\cdot})$  in (\ref{eqn_T2}) by only calculating $\prod_{k\in F_{-}}w_k^{l_1}(\mathbf{x}_{i\cdot})w_k^{l_2}(\mathbf{x}_{i\cdot})$ with $F_{-}=F \cap (F^{l_1} \cup F^{l_2})$,
because $\prod_{k\in F\backslash F_{-}}w_k^{l_1}(\mathbf{x}_{i\cdot})w_k^{l_2}(\mathbf{x}_{i\cdot})=1$.
In combination with the proposition below, computation in (\ref{eqn_T2}) can be dramatically reduced from the full feature set $\mathcal{P}$ to a set only related to the corresponding leaves in a tree.  

\begin{proposition}\label{dimension_reduce}
For any well-defined $p, n, |F|$,
\[
\sum_{k=0}^{p-n}\frac{\binom{p-n}{k}}{p\binom{p-1}{|F|+k}} 
= \frac{1}{n\binom{n-1}{|F|}}.
\]
\end{proposition}

We leave the proof of Proposition \ref{dimension_reduce} in Appendix~\ref{proposition proof}. Further denote $n_{12}=|F^{l_1} \cup F^{l_2}|$ and  a polynomial of $z$,  $P^{l_1l_2}(z)=\prod_{k \in F^{l_1} \cup F^{l_2} \backslash j }(z+w_k^{l_1}(\mathbf{x}_{i\cdot})w_k^{l_2}(\mathbf{x}_{i\cdot})).$
We then define a coefficient polynomial  $C_{n_{12}}(z)=\frac{1}{\binom{n_{12}-1}{0}}z^0 + \frac{1}{\binom{n_{12}-1}{1}}z^1 + \ldots + \frac{1}{\binom{n_{12}-1}{n_{12}-1}}z^{n_{12}-1}.$

\begin{theorem}\label{T2}
The Shapley value in (\ref{msq}) can be calculated as,
\begin{eqnarray}\label{msqdec}
\phi_{\hat{m}^2,j}(\mathbf{x}_{i\cdot}) = \sum_{l}T_{j}^{ll}(\mathbf{x}_{i\cdot}) + 2\sum_{l_1 \neq l_2}T_{j}^{l_1l_2}(\mathbf{x}_{i\cdot}),
\end{eqnarray}
with
\begin{eqnarray}\label{nppoly}
T_{j}^{l_1l_2}(\mathbf{x}_{i\cdot}) &=& \frac{1}{n_{12}} (w_j^{l_1}(\mathbf{x}_{i\cdot})w_j^{l_2}(\mathbf{x}_{i\cdot})-1)\hat{m}_{\emptyset}^{l_1}\hat{m}_{\emptyset}^{l_2} \nonumber\\
&& \times [C_{n_{12}}(z) \cdot P^{l_1l_2}(z)].
\end{eqnarray}
\end{theorem}

\textbf{Proof}.  
With Proposition \ref{dimension_reduce}, we can write (\ref{eqn_T2}) as 
\begin{eqnarray*}
\lefteqn{T_{j}^{l_1l_2}(\mathbf{x}_{i\cdot})}\\
&=& \frac{1}{n_{12}} (w_j^{l_1}(\mathbf{x}_{i\cdot})w_j^{l_2}(\mathbf{x}_{i\cdot})-1)\hat{m}_{\emptyset}^{l_1}\hat{m}_{\emptyset}^{l_2} \\
&& \times \sum_{t=0}^{n_{12}-1}\frac{1}{\binom{n_{12}-1}{t}}\sum^{|F|=t}_{F \subseteq F^{l_1}\cup F^{l_2} \backslash j}\prod_{k \in F}w_k^{l_1}(\mathbf{x}_{i\cdot})w_k^{l_2}(\mathbf{x}_{i\cdot}).
\end{eqnarray*}

We notice that $\sum^{|F|=t}_{F \subseteq F^{l_1}\cup F^{l_2} \backslash j}\prod_{k \in F}w_k^{l_1}(\mathbf{x}_{i\cdot})w_k^{l_2}(\mathbf{x}_{i\cdot})$ is the coefficient of $z^t$ in polynomial $P^{l_1l_2}(z)$, hence the equation (\ref{nppoly}) holds with $C_{n_{12}}(z)$ adjusting the weight based on the size of set $F$. The calculation in (\ref{msqdec}) follows (\ref{mseqsplit}). \hfill 
$\blacksquare$

We only need to consider feature $j \in $$|F^{l_1} \cup F^{l_2}|$ as, otherwise, we have $T_{j}^{l_1l_2}(\mathbf{x}_{i\cdot})=0$ following the definition of $w_j^l(\mathbf{x}_{i\cdot})$. Note that, when there is a feature in set $F$ that doesn't belong to $F^{l_1}(\mathbf{x}_{i\cdot}) \cap F^{l_2}(\mathbf{x}_{i\cdot})\backslash j$, we have $\prod_{k \in F}w_k^{l_1}(\mathbf{x}_{i\cdot})w_k^{l_2}(\mathbf{x}_{i\cdot})=0.$
Thus we can further simplify the term to  
\begin{eqnarray*}
\lefteqn{T_{j}^{l_1l_2}(\mathbf{x}_{i\cdot})}\\
&=& \frac{1}{n_{12}} (w_j^{l_1}(\mathbf{x}_{i\cdot})w_j^{l_2}(\mathbf{x}_{i\cdot})-1)\hat{m}_{\emptyset}^{l_1}\hat{m}_{\emptyset}^{l_2} \sum_{t=0}^{n_{12}-1}\frac{1}{\binom{n_{12}-1}{t}}\\
&& \times \sum^{|F|=t}_{F \subseteq F^{l_1}(\mathbf{x}_{i\cdot})\cap F^{l_2}(\mathbf{x}_{i\cdot}) \backslash j}\prod_{k \in F}w_k^{l_1}(\mathbf{x}_{i\cdot})w_k^{l_2}(\mathbf{x}_{i\cdot}).
\end{eqnarray*}
Consequently, the evaluation of $P^{l_1l_2}(z)$ can be reduced to a much smaller set.

\subsection{The Algorithm}\label{inner}

In this section, we will introduce a fast and stable evaluation for the dot product of a coefficient polynomial $C(z)$ where we know the coefficients and a polynomial $P(z)$ with a known product form, involved in Theorem \ref{T2}.

\begin{proposition}\label{product}
Let $\omega$ be a vector of the complex $n$-th roots of unity whose element is $\exp(\frac{2k\pi i}{n})$ for $k=0, 1, \ldots, n-1$, c the coefficient vector of $C(z)$, and IFFT the Inverse Fast Fourier Transformation. Then
\[
C(z) \cdot P(z) = P(\omega)^T \text{IFFT}(c).
\]
\end{proposition}

The proof of Proposition \ref{product} is shown in Appendix~\ref{proposition proof}. We facilitate the computation via the complex roots of unity because of their numerical stability and fast operations in matrix multiplications. Due to the potential issue of ill condition, especially at large degrees, our calculation avoids inversion of the Vandermonde matrices, although it has been proposed to facilitate the computing by \citet{bifet2022linear}. In addition, for each sample size $n$, we only need to calculate IFFT$(c)$ once, up to order $D$ in $O(n \log(n))$ operations, and the results can be saved for the rest of calculation through Q-SHAP. Note that term $k$ and term $n-k$ in $P(w)$ are complex conjugates, and, for a real vector $c$, IFFT$(c)$ also has the conjugate property for paired term $k$ and term $n-k$. Consequently, the dot product of $P(\omega)$ and IFFT$(c)$ inherits the conjugate property and its imaginary parts are canceled upon addition. Therefore, we only need evaluate the dot product at half of the $n$ complex roots. 

We can aggregate the values of leaf combinations to derive the Shapley values of squared predictions using Q-SHAP as in Algorithm \ref{q-shap} and then calculate the Shapley values of $R^2$ using RSQ-SHAP as in Algorithm \ref{rsq-shap}. The calculation of feature-specific $R^2$ uses the iterative Algorithm \ref{q-shap} instead of a recursive one. As detailed in Appendix~\ref{complexity_algo}, the time complexity of the algorithm is $O(L^2D^2)$ for a single tree,  which doesn't depend on the dimension $p$ and is extremely fast when the maximum tree depth is not too large. 

\begin{algorithm}[!ht]
\caption{\textbf{Q-SHAP}} \label{q-shap}
\begin{algorithmic}
\State \textbf{Q-SHAP}($\mathbf{x}_{i\cdot}$)
\State Initialize $T[j]=0$ for $j=1, \cdots, p$
\For{$l_1$  $\in$ index set ${0, \ldots}, L-1$}
\For {$l_2$ $\in$ index set $l_1, \ldots, L-1$}
    \State Let $n_{12}=|F^{l_1} \cup F^{l_2}|$
    \For {$j \in F^{l_1} \cup F^{l_2}$}
    \State Let  $t[j] =\frac{1}{n_{12}} [w_j^{l_1}(\mathbf{x}_{i\cdot})w_j^{l_2}(\mathbf{x}_{i\cdot})-1] \times$ 
    \State $\hspace{1.5cm}\hat{m}_{\emptyset}^{l_1}\hat{m}_{\emptyset}^{l_2} [C_{n_{12}}(z) \cdot P^{l_1l_2}(z)]$
    \If{$l_1 \neq l_2$}
    \State $T[j] = T[j] + 2t[j]$
    \Else
    \State
    $T[j] = T[j] + t[j]$
    \EndIf
    \EndFor
\EndFor
\EndFor
\State return $T=(T[1],T[2],\cdots, T[p])$
\end{algorithmic}
\end{algorithm}

\begin{algorithm}[!ht]
\caption{\textbf{RSQ-SHAP}} \label{rsq-shap}
\begin{algorithmic}
\State \textbf{RSQ-SHAP}($j$) = $-\frac{1}{Q_{\emptyset}} \Sigma_{i=1}^n \{\textbf{Q-SHAP}(\mathbf{x}_{i\cdot})[j]$ 
\State $\hspace{3.9cm} - 2 y_i \textbf{SHAP}(\mathbf{x}_{i\cdot})[j] \}$
\end{algorithmic}
\end{algorithm}

\section{The Algorithm Q-SHAP for Tree Ensembles from Boosting}\label{qshap-booting}

Tree ensembles from Gradient Boosted Machines (GBM) \citep{friedman2001greedy} greatly improve predictive performance by aggregating many weak learners \citep{chen2016xgboost, ke2017lightgbm, prokhorenkova2018catboost}. Each tree, say tree $k$, is constructed on the residuals from the previous tree, i.e., tree $k-1$. We assume that there are a total of $K$ trees in the ensemble and the quadratic loss by the first $k$ trees, with all features in $\mathcal{P}$, is $Q_{\mathcal{P}}^{(k)}$. Denoting $Q_{\mathcal{P}}^{(0)} = Q_{\emptyset}$, the $k$-th tree reduces the loss by 
\begin{eqnarray}\label{btdiff}
\Delta Q_{\mathcal{P}}^{(k)} = Q_{\mathcal{P}}^{(k-1)} - Q_{\mathcal{P}}^{(k)},
\end{eqnarray}
with the tree ensemble reducing the total loss by
\[
Q_{\emptyset}-Q_{\mathcal{P}}^{(K)} = \sum_{k=1}^K \Delta Q_{\mathcal{P}}^{(k)}.
\]
Per our interest in feature-specific $R^2$,  we resort to the quadratic loss defined as the sum of squared errors in (\ref{eqn_qloss}).

On the other hand, the $k$-th tree provides the prediction $\hat{m}_{\mathcal{P}}^{(k)}(\mathbf{x}_{i\cdot})$. Therefore, the prediction by the first $k$ trees can be recursively calculated as $\hat{y_i}^{(k)}(\mathbf{X}) =\hat{y}_i^{(k-1)}(\mathbf{X})+\alpha\hat{m}_{\mathcal{P}}^{(k)}(\mathbf{x}_{i\cdot}), $where $\alpha$ is the learning rate and $\hat{y}_i^{(0)}(\mathbf{X})\equiv \bar{y}$. Note that the residuals after building $(k-1)$ tree are $\{r_i^{(k-1)}=y_i-\hat{y}^{(k-1)}_i(\mathbf{X}): i=1, 2, \cdots, n\}, $
which are taken to build the $k$-th tree. Thus,
\begin{eqnarray*}
\Delta Q_{\mathcal{P}}^{(k)} &=& \sum_{i=1}^n(r_i^{(k-1)})^2-\sum_{i=1}^n(r_i^{(k-1)} -  \alpha\hat{m}^{(k)}_{\mathcal{P}}(\mathbf{x}_{i\cdot}))^2 \\ 
&=& -\sum_{i=1}^n(\alpha^2 \hat{m}^{(k)2}_{\mathcal{P}}(\mathbf{x}_{i\cdot})-2\alpha r_i^{(k-1)} \hat{m}^{(k)}_{\mathcal{P}}(\mathbf{x}_{i\cdot})).
\end{eqnarray*}

Thus, decomposition of $\Delta Q_{\mathcal{P}}^{(k)}$ in (\ref{btdiff}) for feature-specific Shapley values can be conducted via the decomposition of two sets of values, i.e., SHAP on the predicted value $\hat{m}_{\mathcal{P}}^{(k)}(\mathbf{x}_{i\cdot})$ and Q-SHAP on its quadratic term $\hat{m}_{\mathcal{P}}^{(k)2}(\mathbf{x}_{i\cdot})$. Summing up these Shapley values over all trees leads to Shapley values for the tree ensemble.

\section{Simulation Study} \label{simulation_study}

One of the challenges in assessing methods that explain predictions is the typical absence of a definitive ground truth. Therefore, to fairly demonstrate the fidelity of our methodology,  we must rely on synthetic data that allows for the calculation of the theoretical Shapley values. Here we consider three different models,
\begin{eqnarray*}
&a.& Y = 4 X_1 -5 X_2 + 6 X_3 + \epsilon; \\
&b.& Y = 4 X_1 -5 X_2 + 6 X_3 + 3 X_1 X_2 - X_1 X_3 + \epsilon;\\
&c.& Y = 4 X_1 -5 X_2 + 6 X_3 + 3 X_1 X_2 - X_1 X_2 X_3 + \epsilon. \ \ \ \ \ \ \ \ \ \ \ \ \ \ \ \ \ \ \ \ \ \ \ \ \ \ \ \ \ \ \ \ \ \ \ \ \ \ \ \ \ \ \ \ \ \ \ \ \ \ \ \ \ \ \ \ 
\end{eqnarray*}
All three features involved in the models are generated from Bernoulli distributions with probabilities of 0.6, 0.7, and 0.5, respectively.

We also simulate additional nuisance features independently from $Bernoulli(0.5)$ to make the total number of features $p=100$ and $p=500$, respectively. The error term $\epsilon$ is generated from $N(0, \sigma_{\epsilon}^2)$ with $\sigma_{\epsilon}$ at three different levels, i.e., $0.5$, $1$, and $1.5$. The theoretical values of total $R^2$ and feature-specific $R^2$ are shown in Table~\ref{table-theoreticalrsq} of Appendix~\ref{calcrseq}.

\begin{table*}[ht]
\centering
 \caption{Estimation bias (SE) of $X_1$-specific, $X_2$-specific, $X_3$-specific, and the sum of all feature-specific $R^2$ values across the three models with $n=1,000$, $p=100$, and $\sigma_{\epsilon}=1.5$. Bolded values indicate the smallest bias in magnitude.} \label{Table_SimuBias}
\begin{tabular}{llcccccc}
\toprule
 & Method & $X_1$-specific $R^2$ & $X_2$-specific $R^2$ & $X_3$-specific $R^2$ & Sum of all $R^2$ \\
\midrule
\multirow{3}{*}{Model a} 
     & Q-SHAP & \textbf{0.006} (0.011) & \textbf{0.008} (0.014) & \textbf{0.015} (0.015) & \textbf{0.031} (0.014) \\
    & SAGE   & -0.015 (0.015) & -0.017 (0.016) & -0.021 (0.016) & -0.053 (0.020) \\
    & SPVIM  & 0.049 (0.035) & 0.069 (0.039) & 0.116 (0.051) & 0.242 (0.208) \\
\midrule
\multirow{3}{*}{Model b} 
    & Q-SHAP & \textbf{0.008} (0.017) & \textbf{0.002} (0.009) & \textbf{0.009} (0.016) & \textbf{0.024} (0.019) \\
    & SAGE   & -0.026 (0.016) & -0.014 (0.011) & -0.024 (0.014) & -0.062 (0.019) \\
    & SPVIM  & 0.111 (0.046) & 0.049 (0.033) & 0.098 (0.048) & 0.256 (0.219) \\
\midrule
\multirow{3}{*}{Model c} 
    & Q-SHAP & \textbf{0.005} (0.017) & \textbf{0.002} (0.009) & \textbf{0.005} (0.015) & \textbf{0.021} (0.016) \\
    & SAGE   & -0.025 (0.016) & -0.014 (0.012) & -0.024 (0.014) & -0.062 (0.020) \\
    & SPVIM  & 0.107 (0.047) & 0.048 (0.032) & 0.098 (0.046) & 0.260 (0.206) \\\bottomrule
\end{tabular}
\end{table*}

We evaluate the performance of three different methods, our proposed Q-SHAP, SAGE by \citet{covert2020understanding}, and SPVIM by \citet{williamson2020efficient}, in calculating the feature-specific $R^2$ for the above three models with data sets of different sample sizes at $n=500, 1000, 2000$, and $5000$. We use package \textit{sage-importance} for SAGE and 
package \textit{vimpy} for SPVIM to calculate feature-specific Shapley values of total explained variance, which are divided by the total variance for corresponding feature-specific $R^2$ values.

For each setting, we generated 1,000 data sets. For each data set, we built a tree ensemble using XGBoost \citep{chen2016xgboost} with tuning parameters optimized via 5-fold cross-validation and grid search in a parameter space specified with the learning rate in $\{0.01, 0.05, 0.1\}$ and number of estimators in $\{50, 100, 200, 300, \cdots, 1000\}$. We fixed the maximum depth of models a, b, and c at 1, 2, and 3 respectively. Table~\ref{Table_SimuBias} shows the bias in calculating feature-specific $R^2$ for the first three features as well as the sum of all feature-specific $R^2$ for all three models with $n=1000$, $p=100$, and $\sigma_{\epsilon}=1.5$. The estimation results of the three models in other settings are plotted in Appendix~\ref{simulation_supp}. Overall, Q-SHAP provides a more stable and accurate calculation of feature-specific $R^2$ than the other two methods.

\begin{figure}[htb]
    \centering
    \includegraphics[width=0.485\textwidth]{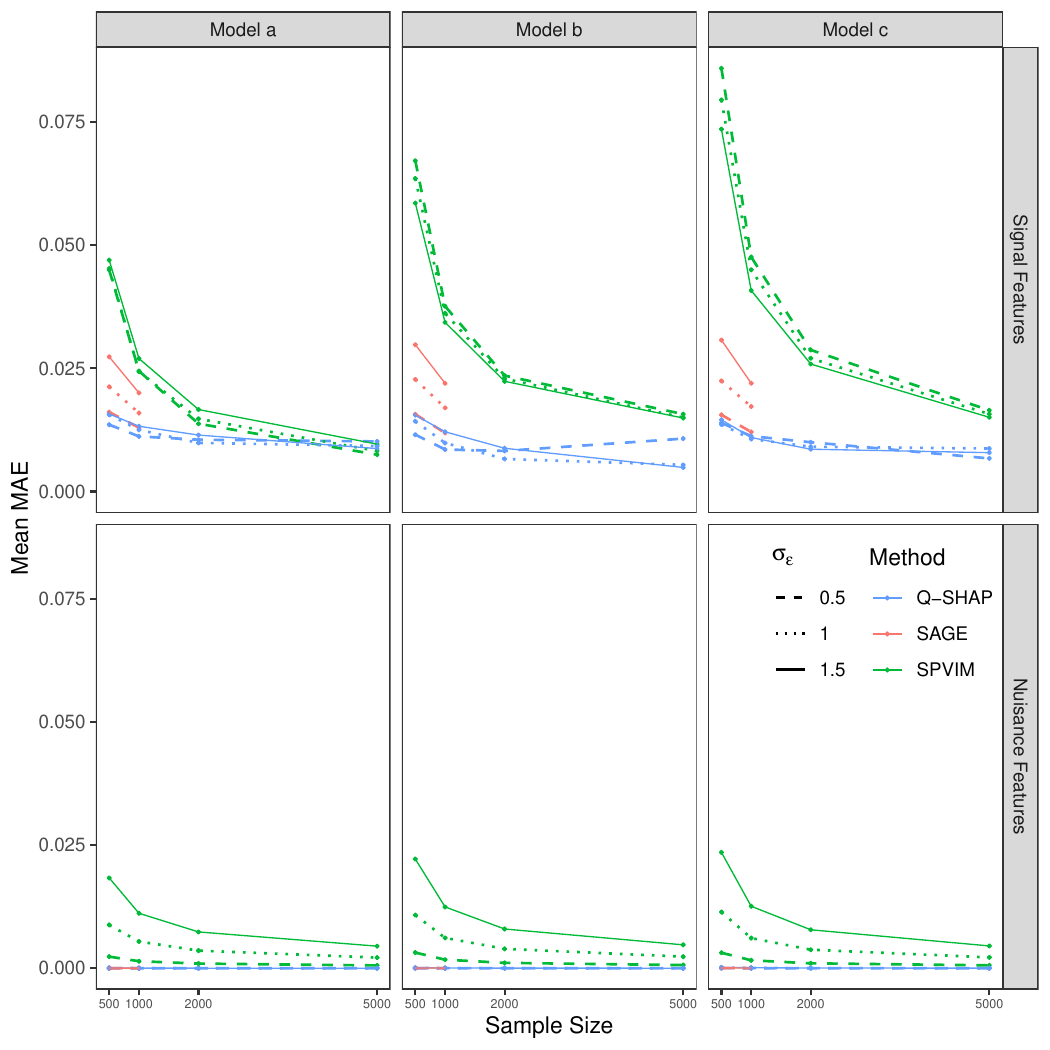} 
    \caption{The mean absolute error (MAE) averaged over 1,000 datasets with $p=100$} \label{mmae}
\end{figure}

We divide all features into two groups, signal features (the first three) and nuisance features (the rest). For each group, we calculated the mean absolute error (MAE) by comparing feature-specific $R^2$ values to the theoretical ones in each of the 1,000 datasets and averaged MAE over the 1,000 datasets, shown in Figure~\ref{mmae}. Note that, by limiting memory to 2GB, SAGE can only report $R^2$ for the data sets with sample size at 500 and 1,000.

For both signal and nuisance features, Q-SHAP and SAGE exhibit consistent behavior across all models. In contrast, SPVIM tends to bias the calculation, especially for small sample sizes, indicated by the rapid increase of MMAE when the sample size goes down. Among signal features, Q-SHAP has better accuracy than SAGE, followed by SPVIM in general. All methods tend to have better accuracy when sample size increases. 

For the nuisance features, only SPVIM is biased away from 0. On the other hand, both Q-SHAP and SAGE have almost no bias for nuisance features across different sample sizes. For all three methods, $R^2$ of signal features tends to have a larger bias than nuisance features. 

\begin{figure}[htb]
    \centering 
    \includegraphics[width=0.485\textwidth]{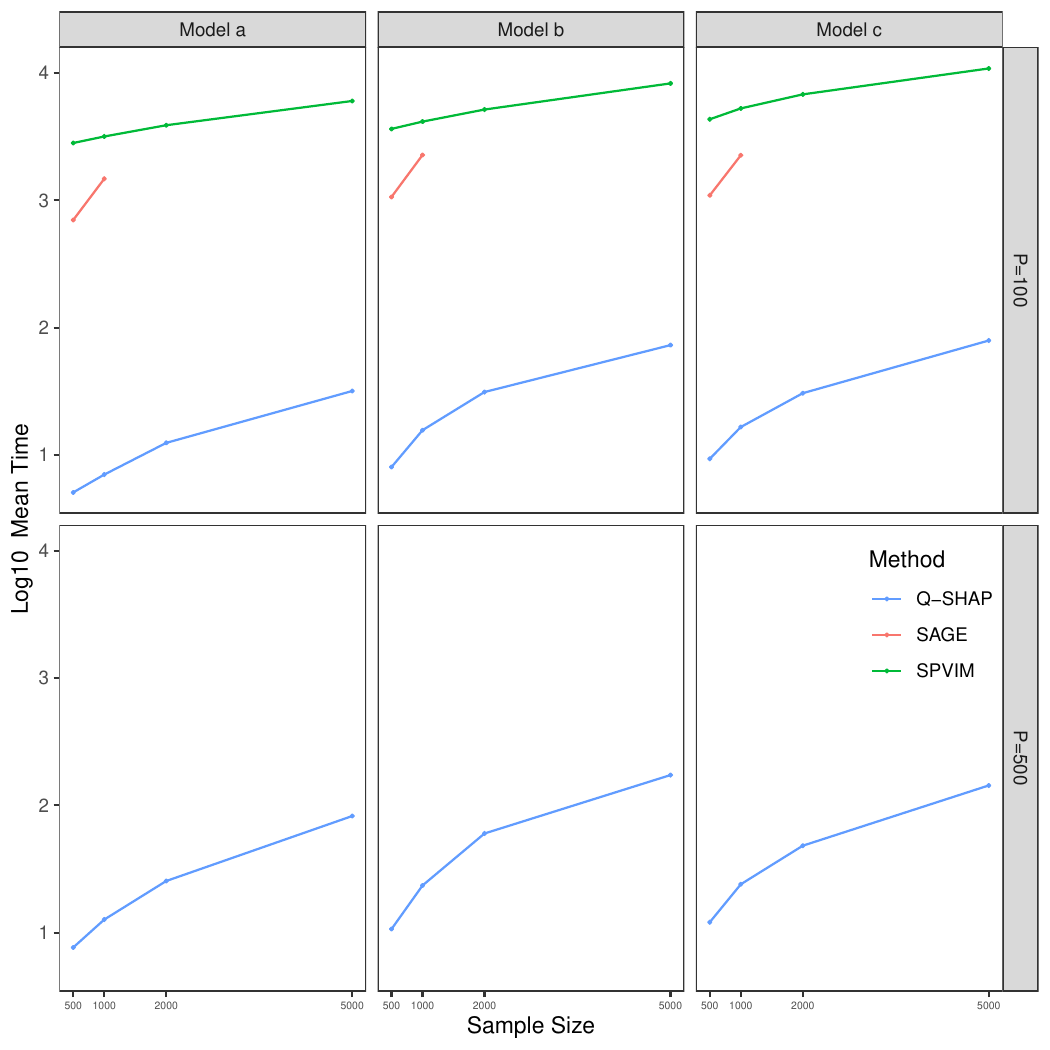}
    \caption{The running time (seconds) in simulation study} \label{running_time}
\end{figure}

We compared the computational time of the three different methods by running all algorithms in parallel on a full node consisting of two AMD CPUs@2.2GHz with 128 cores and 256 GB memory. We unified the environment with the help of a Singularity container \citep{kurtzer2017singularity} built under Python version 3.11.6. Due to the large size of the simulation, we limit all methods to a maximum wall time of 4 hours per dataset on a single core, with memory limited to 2 GB. The running times are shown in Figure~\ref{running_time}. Both SAGE and SPVIM demanded a long time to compute even with only 100 features. Q-SHAP is hundreds of times faster than both SAGE and SPVIM in general and is the only method that can be completed when the dimension is 500 in constrained computation time and memory.

\section{Real Data Analysis} \label{real}

We illustrate the utility of Q-SHAP by applying it to three datasets: (1) \underline{Healthcare} data that includes eight features for each of the 1338 subjects besides their healthcare insurance costs \footnote{\url{https://www.kaggle.com/datasets/arunjangir245/healthcare-insurance-expenses/data}}; (2) \underline{Prostate} data that includes expression levels of 17,261 genes for 551 samples, available from UCSC Xena \citep{goldman2020visualizing}, and cancer-indicating Gleason score, available from TCGAbiolinks \citep{colaprico2016tcgabiolinks};
and (3) \underline{S\&P 500} data that includes prices of NVIDIA and other 469 stocks for 1,258 business days from February 8, 2013 to February 7, 2018 \footnote{\url{https://www.kaggle.com/datasets/camnugent/sandp500/data}}. Here we predicted the daily return rate of NVIDIA stock from those of other 469 stocks in S\&P 500 data, and the Gleason score, adjusted for age and race, using the gene expression features in Prostate data. 

For each data, we constructed the tree ensemble using XGBoost with tuning parameters optimized via 5-fold cross-validation and random search in a parameter space specified with the number of trees in $\{50, 100, 500, 1000, 1500, 2000, 2500, 3000\}$, maximum depth in $\{1,2,\cdots, 6\}$, and learning rate in $\{0.01, 0.05, 0.1\}$. For Prostate and S\& P 500 data, as both SAGE and SPVIM cannot manage the large numbers of features, we first applied Q-SHAP to the tree ensembles built on all features and then selected the top 100 features to reconstruct the tree ensembles, whose feature-specific $R^2$ were calculated using all three methods. As shown in Table~\ref{realdata_computing_time}, although feasible for limited number of features, both SAGE and, in particular, SPVIM took much longer time to calculate the feature-specific $R^2$ values. 

\begin{table}[htbp]
\centering
\caption{The running time in real data analysis}\label{realdata_computing_time}
\begin{tabular}{lccc}
\toprule
& \textbf{Q-SHAP} & \textbf{SAGE} & \textbf{SPVIM} \\
\midrule
\textbf{Healthcare} & 32 s     & 6 m     & 22 m      \\
\textbf{Prostate} & 128 s     & 43 m    &  67 h   \\
\textbf{S\&P 500} & 41 s     & 26 h      & 138 h       \\
\bottomrule
\end{tabular}
\end{table}

For the healthcare data, the tree ensemble in predicting healthcare insurance expenses reports the total $R^2$ at 0.86. As shown in Table~\ref{healthcare_table}, both Q-SHAP and SAGE ranked the eight features similarly, although SAGE tends to report slightly smaller values. On the other hand, SPVIM differs from the other two methods with its ranking list, demonstrated by the highlighted features in Table \ref{healthcare_table}. In fact, SPVIM reported some much larger feature-specific $R^2$ values with some much smaller ones which were even negative.

With 100 features in the rebuilt tree ensembles for the prostate and S\&P 500 data, we observe inconsistencies between Q-SHAP and SAGE, while the feature ranking from Q-SHAP matched that of models built on full feature set, see Table~\ref{prostate_table} and Table~\ref{snp500_table}. The rebuilt tree ensembles reported total $R^2$ of $0.995$ for the prostate data and 0.733 for the S\&P 500 data. Notably, for the prostate data, the sum of all 100 feature-specific $R^2$ values from Q-SHAP matches the total $R^2$ value, whereas SAGE yields a slightly lower sum of 0.938. In contrast, although SPVIM tends to overstate some feature-specific $R^2$ values, the total sum of its feature-specific $R^2$ is only -0.22, substantially lower than the total value at 0.995. Overall, the real data analysis aligns with the simulation study, confirming that SAGE tends to underestimate the feature-specific $R^2$, SPVIM shows instability, and Q-SHAP outperforms both in computational efficiency and the accuracy of feature-specific $R^2$. 

\begin{table}[ht]
\centering
\caption{Feature-specific $R^2$ in the healthcare data. Highlighted are features with $R^2$ larger than the preceding one.} \label{healthcare_table}
\begin{adjustbox}{max width=\textwidth}

\begin{tabular}{lccc}
\toprule
\textbf{Feature} & \textbf{Q-SHAP} & \textbf{SAGE} & \textbf{SPVIM} \\
\midrule
smoker\_yes & 0.481 & 0.455 & 0.565 \\
age & 0.323 & 0.293 & 0.327 \\
children  & 0.026 & 0.021 & 0.012 \\
bmi & 0.021 & 0.019 & \textcolor{red}{0.051} \\
sex\_male & 0.003 & 0.000 & -0.033 \\
region\_southwest & 0.002 & \textcolor{red}{0.001} & \textcolor{red}{0.044} \\
region\_southeast & 0.001 & 0.000 & \textcolor{red}{0.048} \\
region\_northwest & 0.000 & 0.000 & -0.020 \\
\bottomrule
\end{tabular}
\end{adjustbox}
\end{table}

\begin{table}[ht]
\centering
\caption{Feature specific $R^2$ in the prostate data. Highlighted are features with $R^2$ larger than the preceding one.}\label{prostate_table}

\begin{adjustbox}{max width=\textwidth}

\begin{tabular}{lcccc}
\toprule
& \underline{Orginal} & \multicolumn{3}{c}{\underline{Model with Selected Feaures}}
\\ 
\textbf{Gene} & \textbf{Q-SHAP} & \textbf{Q-SHAP} & \textbf{SAGE} & \textbf{SPVIM} \\
\midrule
SLC7A4    & 0.105 & 0.112 & 0.063 & 0.031  \\
FAM72A    & 0.051 & 0.063 & 0.045 & -0.025 \\
CENPA     & 0.039 & 0.039 & 0.024 & -0.175 \\
KIAA0319L & 0.034 & 0.036 & \textcolor{red}{0.029} & \textcolor{red}{0.033}  \\
CBX2      & 0.028 & 0.034 & 0.025 & \textcolor{red}{0.063}  \\
COL5A2    & 0.018 & 0.022 & 0.022 & -0.033 \\
SPATA4    & 0.018 & 0.019 & 0.021 &\textcolor{red}{-0.004} \\
DOCK6     & 0.016 & 0.017 & 0.010 & -0.150 \\
KIF18B    & 0.015 & \textcolor{red}{0.023} & \textcolor{red}{0.012} & \textcolor{red}{0.185}  \\
TSEN15    & 0.014 & 0.015 & \textcolor{red}{0.021} & -0.004 \\ \hline
OTHERS    & 0.572 & 0.614 & 0.667 & -0.140 \\
\bottomrule
\end{tabular}
\end{adjustbox}
\end{table}

\begin{table}[ht]
\centering
\caption{Feature-specific $R^2$ in the S\&P 500 data. Highlighted are features with $R^2$ larger than the preceding one.}\label{snp500_table}
\begin{adjustbox}{max width=\textwidth}

\begin{tabular}{lcccc}
\toprule
& \underline{Orginal} & \multicolumn{3}{c}{\underline{Model with Selected Feaures}}
\\ 
\textbf{Stock} & \textbf{Q-SHAP} & \textbf{Q-SHAP} & \textbf{SAGE} & \textbf{SPVIM} \\
\midrule
AMAT   & 0.100   & 0.111 & 0.075 & -0.141 \\
AMD    & 0.077 & 0.085 & 0.071 & \textcolor{red}{0.102}  \\
MCHP   & 0.072 & 0.069 & 0.034 & \textcolor{red}{0.158}  \\
ADI    & 0.068 & \textcolor{red}{0.072} & \textcolor{red}{0.048} & -0.026 \\
TXN    & 0.041 & 0.043 & 0.029 & -0.142 \\
MU     & 0.035 & 0.041 & \textcolor{red}{0.039} & -0.232 \\
ADM    & 0.022 & 0.026 & 0.013 & \textcolor{red}{-0.004} \\
AGN    & 0.021 & 0.023 & \textcolor{red}{0.018} & \textcolor{red}{0.120}   \\
NEM    & 0.016 & 0.015 & 0.010  & \textcolor{red}{0.202}  \\
PBCT   & 0.015 & 0.013 & \textcolor{red}{0.011} & -0.137 \\ \hline
Others & 0.157 & 0.234 & 0.153 & -0.565 \\
\bottomrule
\end{tabular}
\end{adjustbox}
\end{table}

\section{Conclusion}\label{conclusion}

The coefficient of determination, aka $R^2$, measures the proportion of the total variation explained by available features. Its additive decomposition, following \citet{shapley1953value}, provides an ideal evaluation of each feature's attribute to explain the total variation. Shapley values are defined by differences involving all feature subsets, a seemingly  NP-hard problem in general. Recently, several methods 
\citep{lundberg2018consistent, yang2021fast, bifet2022linear} have been developed to leverage the structure of tree-based models and provide computationally efficient algorithms to decompose the predicted values. However, decomposing $R^2$ demands the decomposition of a quadratic loss reduction by multiple trees. We have shown in Section~{\ref{qshap-booting}} that we can attribute the total loss reduction by the tree ensemble to each single tree, and the tree-specific loss reductions are subject to further decomposition to each feature. However, decomposing a quadratic loss of a single tree needs work with the squared terms of predicted values, invalidating previously developed methods for predicted values. Leveraging structural property of trees and theoretical results of polynomials, we developed the Q-SHAP algorithm to consolidate calculations cross models and calculate Shapley values of squared predicted valued in polynomial time. The algorithm works not only for $R^2$ but also for general quadratic losses. Ultimately, it may provide a framework for more general loss functions via approximation.

\begin{acknowledgements}
This research was partially supported by NIH grants R01GM131491, R01AG080917, and R01AG080917-02S1, NCI grants R03 CA235363 and P30CA062203, and UCI Anti-Cancer Challenge funds from the UC Irvine Comprehensive Cancer Center. The results shown in Section~\ref{real} are in part based upon data generated by the TCGA Research Network: https://www.cancer.gov/tcga. The content is solely the responsibility of the authors and does not necessarily represent the official views of the National Institutes of Health or the Chao Family Comprehensive Cancer Center.

\end{acknowledgements}

\bibliography{uai2025-QSHAP}

\newpage
\onecolumn

\title{Fast Calculation of Feature Contributions in Boosting Trees\\(Supplementary Material)}
\maketitle

\appendix

\section{Proofs} \label{proposition proof}

We first establish the following lemma.

\begin{lemma}\label{gould_identity}
\[
\sum_{k=0}^{p-n}\binom{|F|+k}{k}\binom{p-1-|F|-k}{p-n-k} = \binom{p}{n}.
\]
\end{lemma}

\textbf{Proof}. Using Gould's identity \citep{gould1972combinatorial}, 
we have 
\begin{eqnarray*}
\lefteqn{\sum_{k=0}^{p-n}\binom{|F|+k}{k}\binom{p-1-|F|-k}{p-n-k}}\\
&=& \sum_{k=0}^{p-n}\binom{k}{k}\binom{p-k-1}{p-n-k}\\
&=& \sum_{k=0}^{p-n}\binom{p-k-1}{n-1}\\
&=& \binom{p}{n},
\end{eqnarray*}
where we used the Hockey-Stick Identity in the last step. 

\textbf{Proof of Proposition \ref{dimension_reduce}}. Through expansion and Lemma~\ref{gould_identity}, we have
\begin{eqnarray*}
\lefteqn{\sum_{k=0}^{p-n}\frac{\binom{p-n}{k}\binom{n-1}{|F|}}{p\binom{p-1}{|F|+k}}}\\ 
&=&  \frac{1}{p}\sum_{k=0}^{p-n}\frac{(|F|+k)!(p-n)!(p-1-|F|-k)!(n-1)!}{(p-1)!(p-n-k)!k!|F|!(n-1-|F|!)} \\
&=&  \frac{1}{p}\frac{(p-n)!(n-1)!}{(p-1)!}\sum_{k=0}^{p-n}\binom{|F|+k}{k}\binom{p-1-|F|-k}{p-n-k} \\
&=&  \frac{1}{p}\frac{(p-n)!(n-1)!}{(p-1)!}\binom{p-1}{n-1}\frac{p}{n} \\
&=& \frac{1}{n}.
\end{eqnarray*}

\textbf{Proof of Proposition~\ref{product}}. We first rewrite the two polynomials 
\begin{eqnarray*}
\left\{\begin{array}{l}C(z)=V(z)c, \\ P(z)=V(z)a,\end{array}\right.
\end{eqnarray*}
where $V(z)$ is the Vandermonde matrix for vector $z$,  and $c$ and $a$ are the coefficients of polynomials $C(z)$ and $P(z)$, respectively. Then the inner product 
\begin{eqnarray*}
\lefteqn{C(z) \cdot P(z)}\\
&=& P(z) \cdot C(z)\\
&=& a^Tc\\
&=& (V(z)^{-1}P(z))^Tc\\
&=& P(z)^T (V(z)^T)^{-1}c. 
\end{eqnarray*}
Letting 
\[
z=\omega,
\]
and noting that the Vandermonde matrix evaluated at $\omega$ is symmetric, we have 
\[
(V(\omega)^T)^{-1}=V(\omega)^{-1}=\frac{1}{n}V(\omega^{-1}),
\]
whose multiplication with $c$ is just the Inverse Fast Fourier transformation (IFFT) over $c$ \citep{geddes1992algorithms}. Hence the proposition holds.

\section{Theoretical \texorpdfstring{$R^2$}{R-Squared} Values in Simulated Models} \label{calcrseq}

The theoretical total and feature-specific $R^2$ in the three models are shown in Table~\ref{table-theoreticalrsq}.

\begin{table}[H]
\centering\caption{Theoretical $R^2$ in Simulated Models}\label{table-theoreticalrsq}
\begin{tabular}{c|ccccc}
\hline\hline
& & \multicolumn{4}{c}{$R^2$} \\ \cline{3-6}
Model & $\sigma_{\epsilon}$ & Total & $X_1$ & $X_2$ & $X_3$ \\
\hline
 & 0.50 & 0.9864 & 0.2094 & 0.2863 & 0.4907 \\
a & 1.00 & 0.9477 & 0.2012 & 0.2750 & 0.4715 \\
 & 1.50 & 0.8894 & 0.1888 & 0.2581 & 0.4425 \\ \hline
 & 0.50 & 0.9860 & 0.4390 & 0.1341 & 0.4129 \\
b & 1.00 & 0.9459 & 0.4212 & 0.1286 & 0.3961 \\
 & 1.50 & 0.8860 & 0.3945 & 0.1205 & 0.3710 \\ \hline
 & 0.50 & 0.9868 & 0.4288 & 0.1450 & 0.4130 \\
c & 1.00 & 0.9491 & 0.4124 & 0.1395 & 0.3972 \\
 & 1.50 & 0.8925 & 0.3878 & 0.1312 & 0.3735 \\ \hline\hline
\end{tabular}
\end{table}

\section{Additional Simulation Results}\label{simulation_supp}

\subsection{Boxplots of the First Three Feature-Specific and Total \texorpdfstring{$R^2$}{R2} Values} \label{SA_Simu_Rsq}

\begin{figure}[!htb]
    \centering
    \subfigure[$X_1$-specific $R^2$]{
        \includegraphics[width=0.44\textwidth, height=4cm]{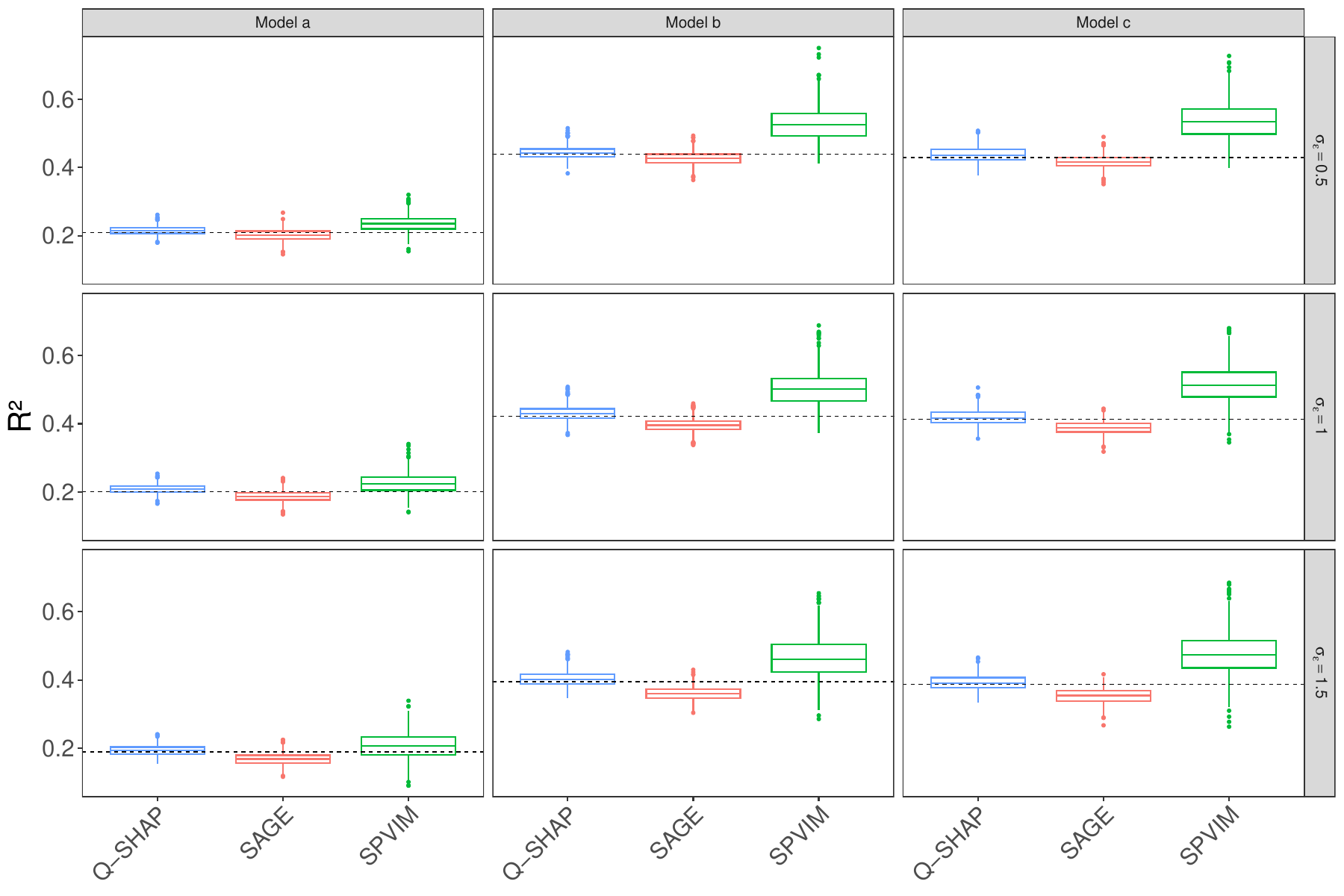}    }
    \subfigure[$X_2$-specific $R^2$]{
        \includegraphics[width=0.44\textwidth, height=4cm]{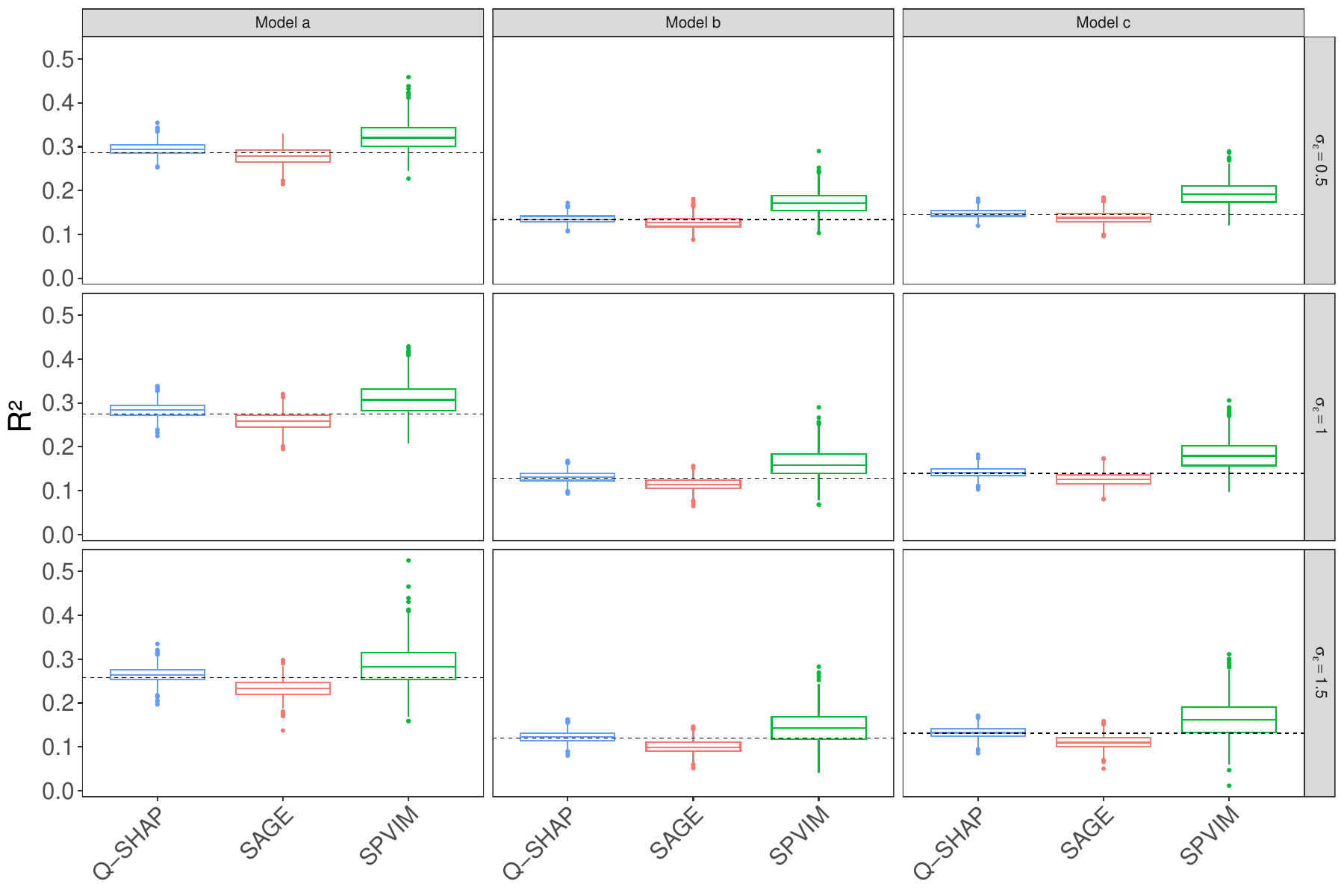}
    }
    
    \subfigure[$X_3$-specific $R^2$]{
        \includegraphics[width=0.44\textwidth, height=4cm]{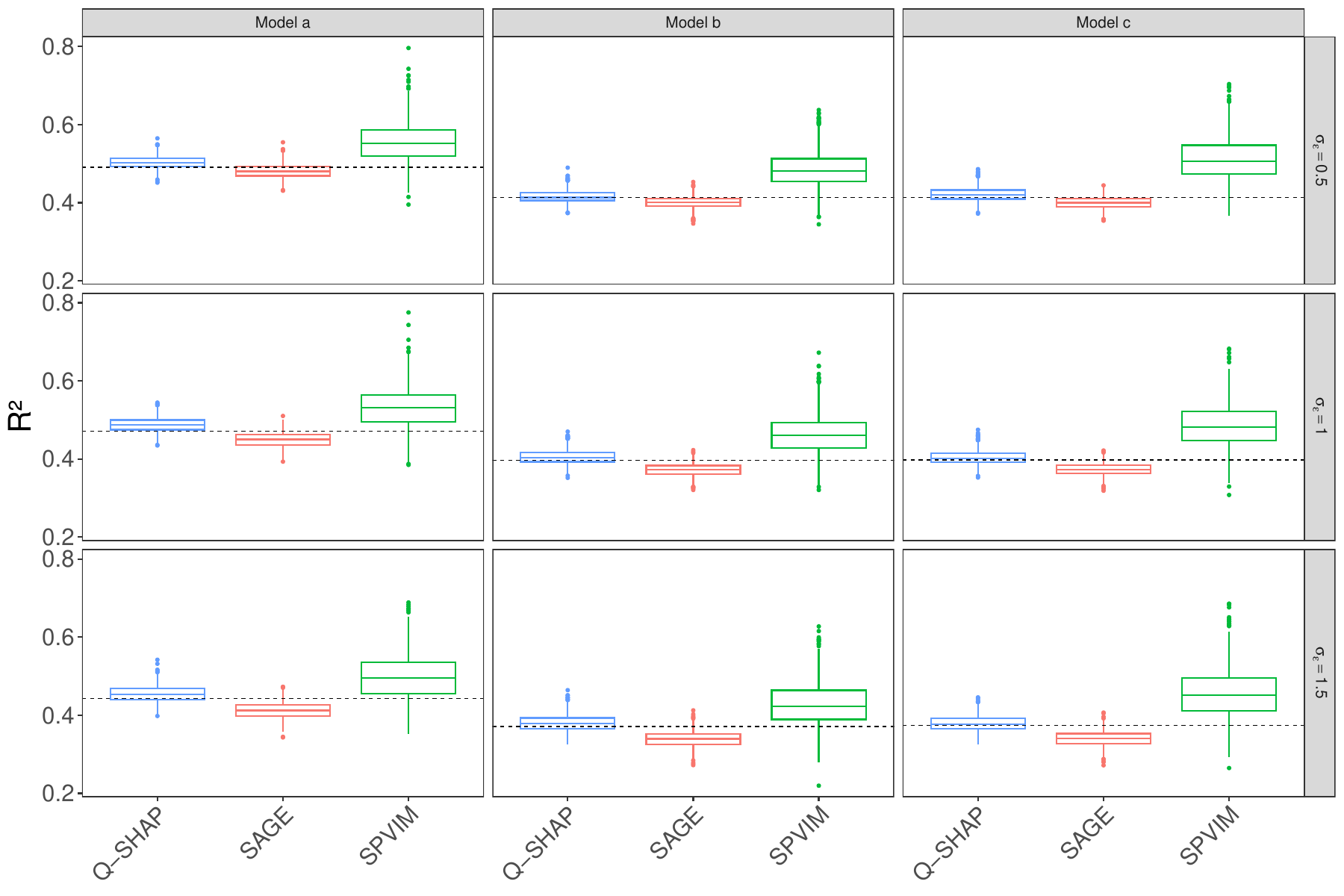}
    }
    \subfigure[Sum of all $R^2$]{
        \includegraphics[width=0.44\textwidth, height=4cm]{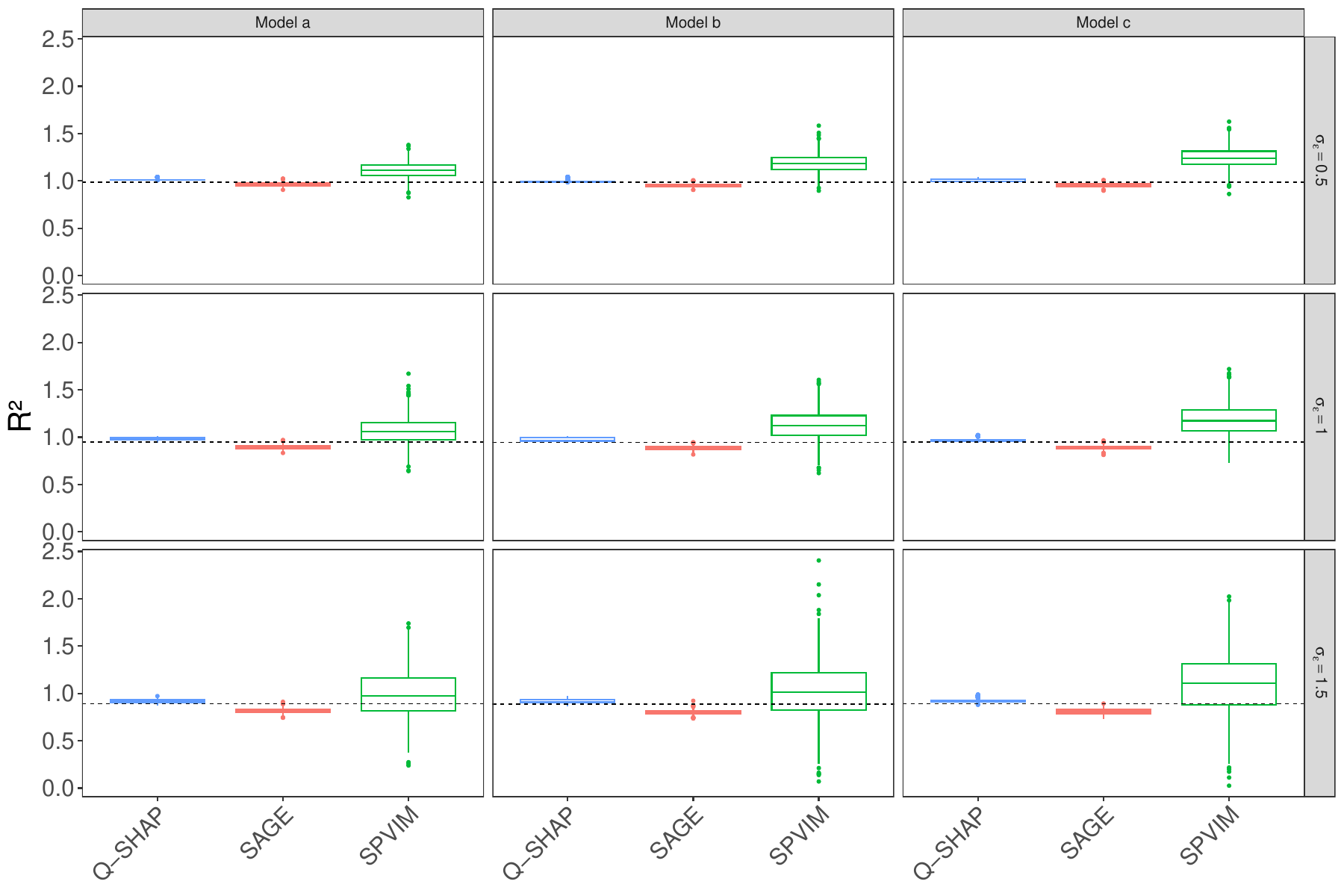}
    }
    \caption{Boxplots of (a) $X_1$-specific, (b) $X_2$-specific, (c) $X_3$-specific, and (d) the sum of all feature-specific $R^2$ in the three models with $n=500$, $p=100$. The dashed lines show the theoretical $R^2$.}\label{np051}
\end{figure}

\begin{figure}[!hbp]
    \centering
    \subfigure[$X_1$-specific $R^2$]{
        \includegraphics[width=0.44\textwidth, height=4cm]{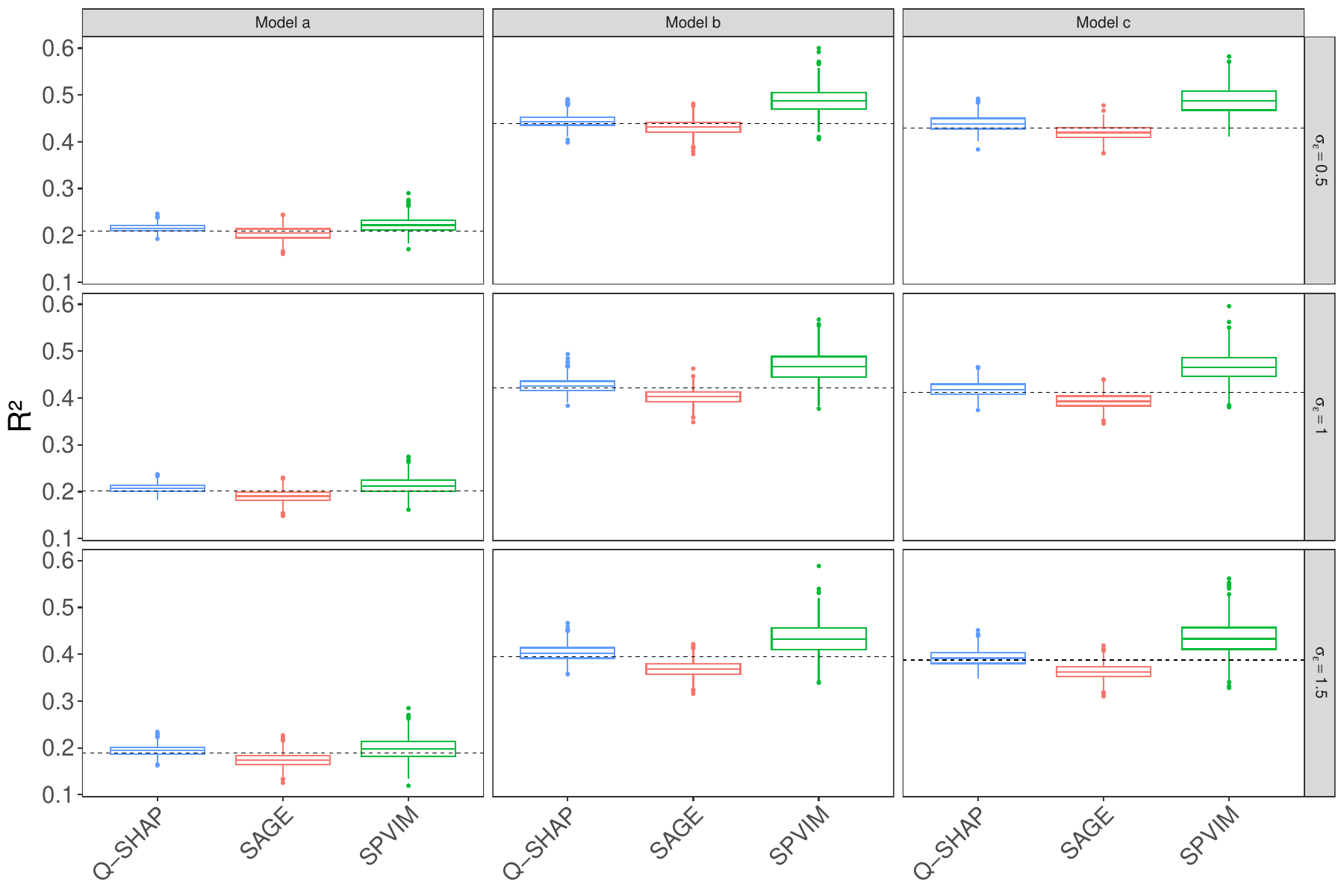}    }
    \subfigure[$X_2$-specific $R^2$]{
        \includegraphics[width=0.44\textwidth, height=4cm]{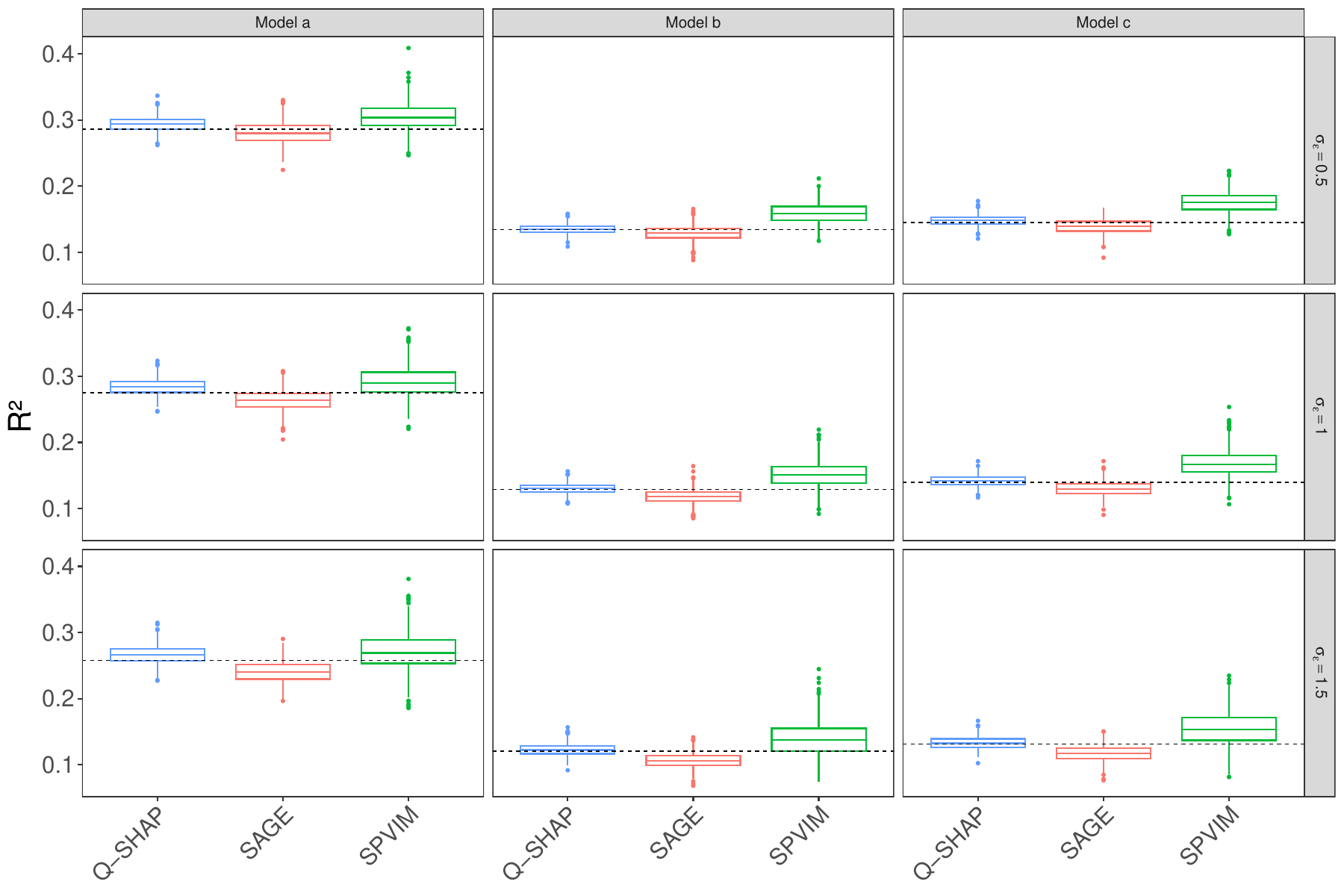}
    }
    
    \subfigure[$X_3$-specific $R^2$]{
        \includegraphics[width=0.44\textwidth, height=4cm]{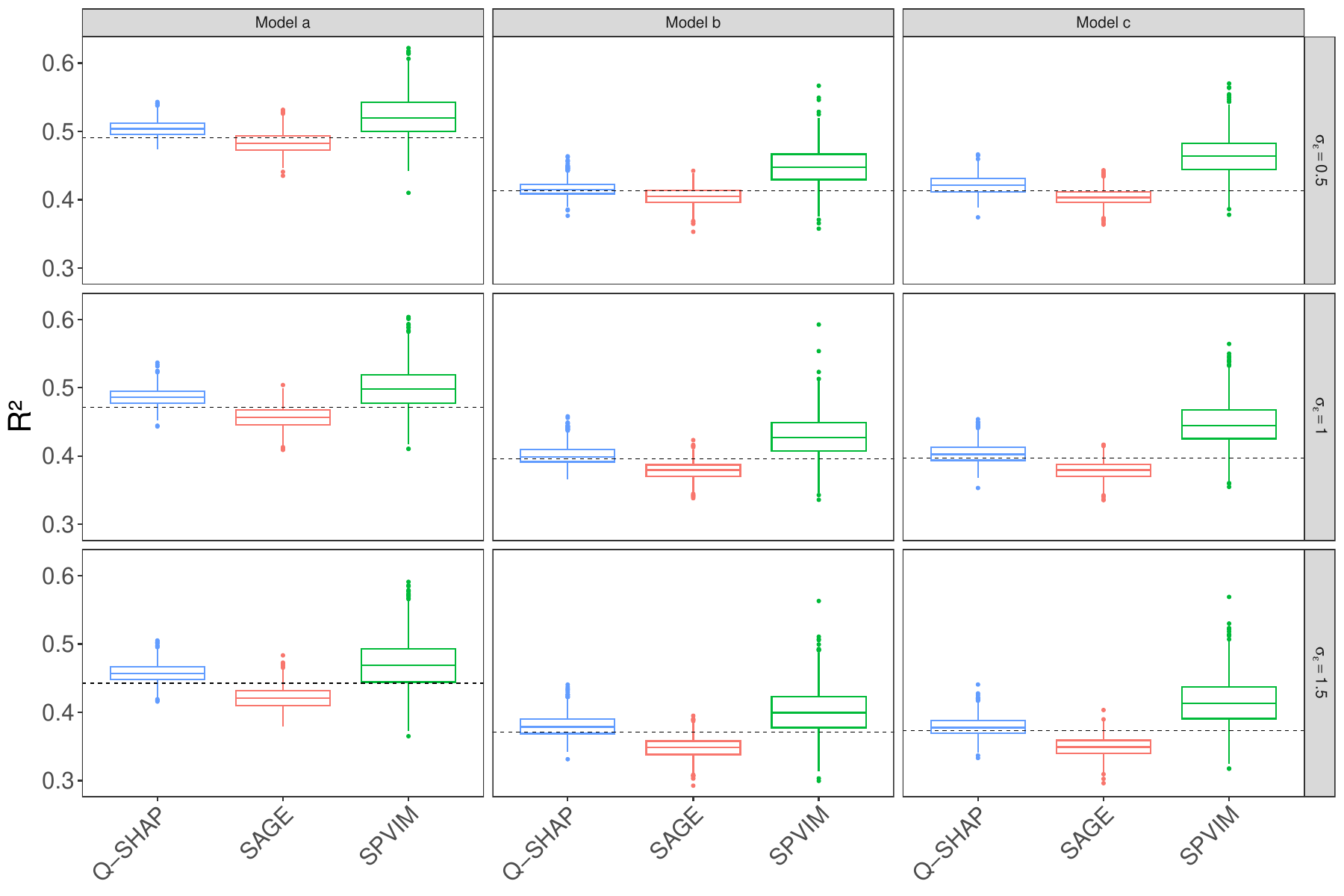}
    }
    \subfigure[Sum of all $R^2$]{
        \includegraphics[width=0.44\textwidth, height=4cm]{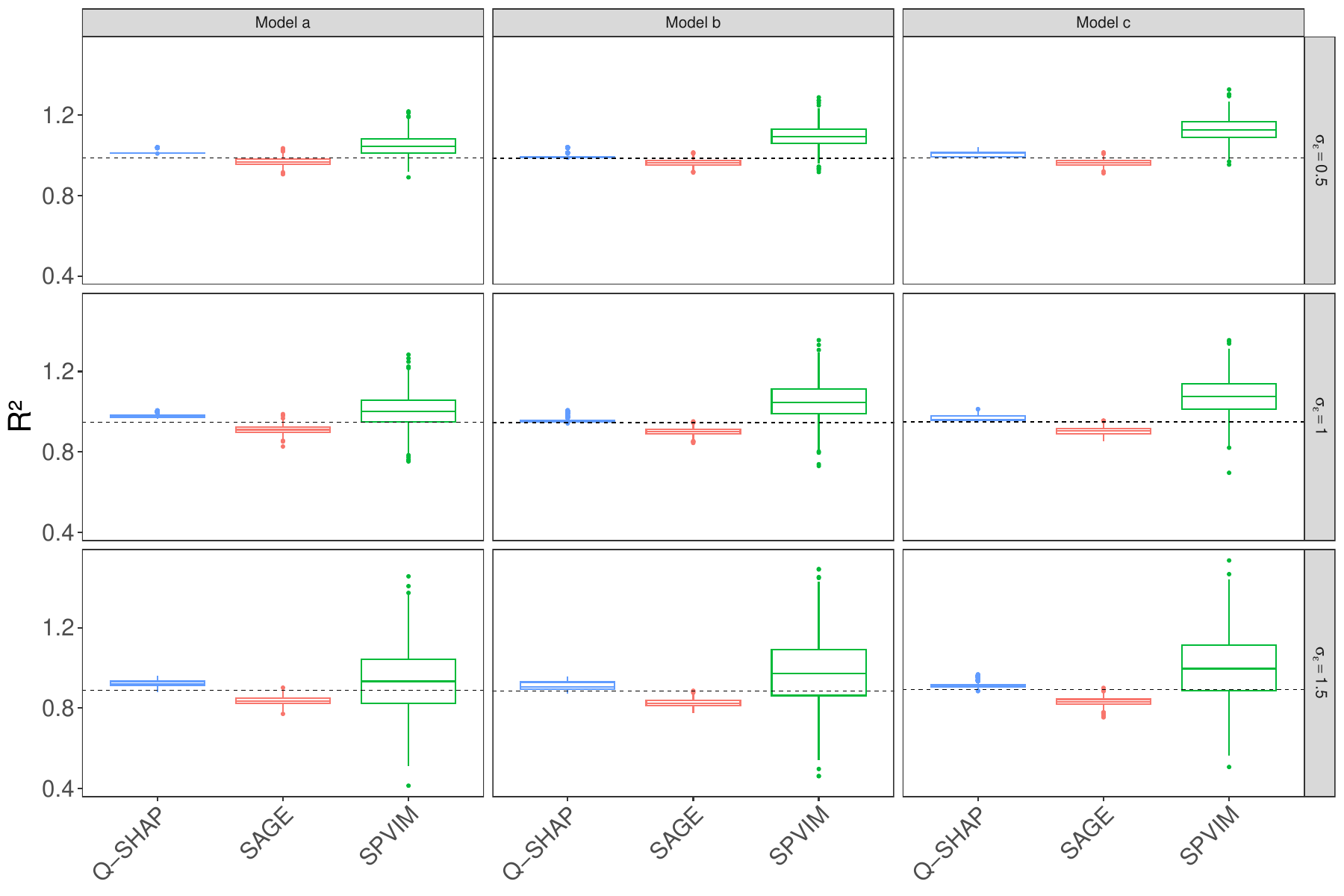}
    }
    \caption{Boxplots of (a) $X_1$-specific, (b) $X_2$-specific, (c) $X_3$-specific, and (d) the sum of all feature-specific $R^2$ in the three models with $n=1000$, $p=100$. The dashed lines show the theoretical $R^2$.}\label{np101}
\end{figure}

\begin{figure}[!hbp]
    \centering
    \subfigure[$X_1$-specific $R^2$]{
        \includegraphics[width=0.44\textwidth, height=4cm]{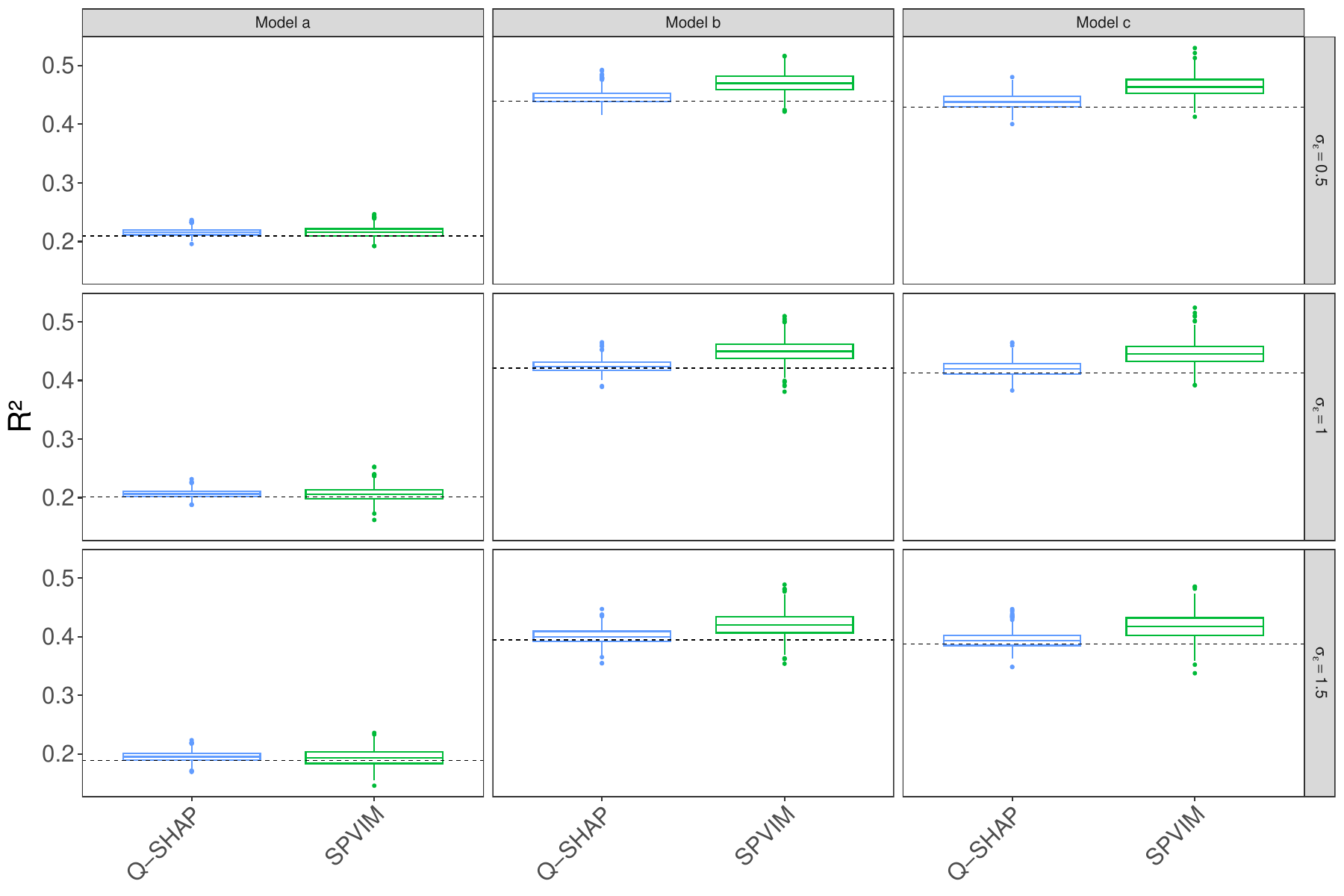}    }
    \subfigure[$X_2$-specific $R^2$]{
        \includegraphics[width=0.44\textwidth, height=4cm]{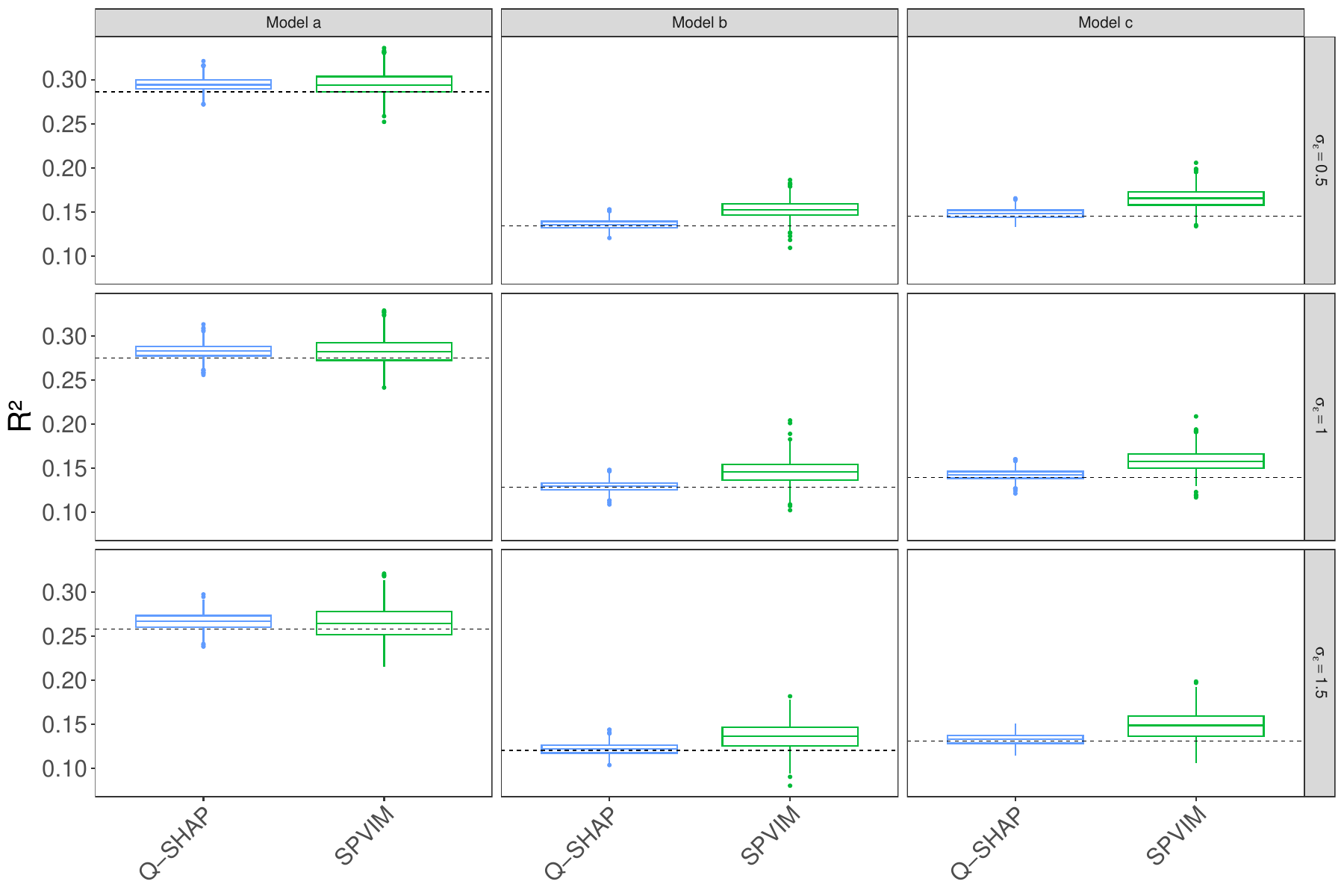}
    }
    
    \subfigure[$X_3$-specific $R^2$]{
        \includegraphics[width=0.44\textwidth, height=4cm]{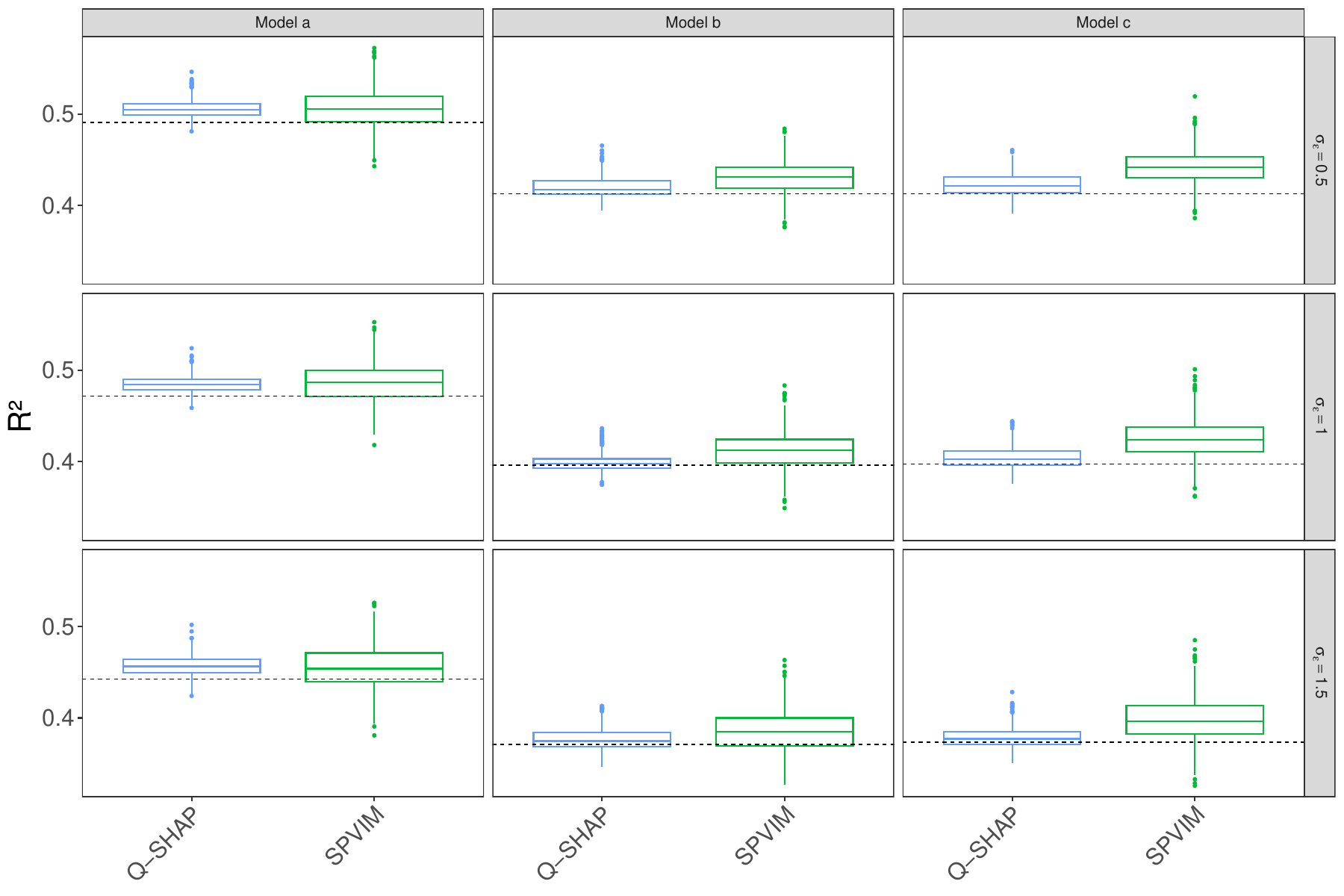}
    }
    \subfigure[Sum of all $R^2$]{
        \includegraphics[width=0.44\textwidth, height=4cm]{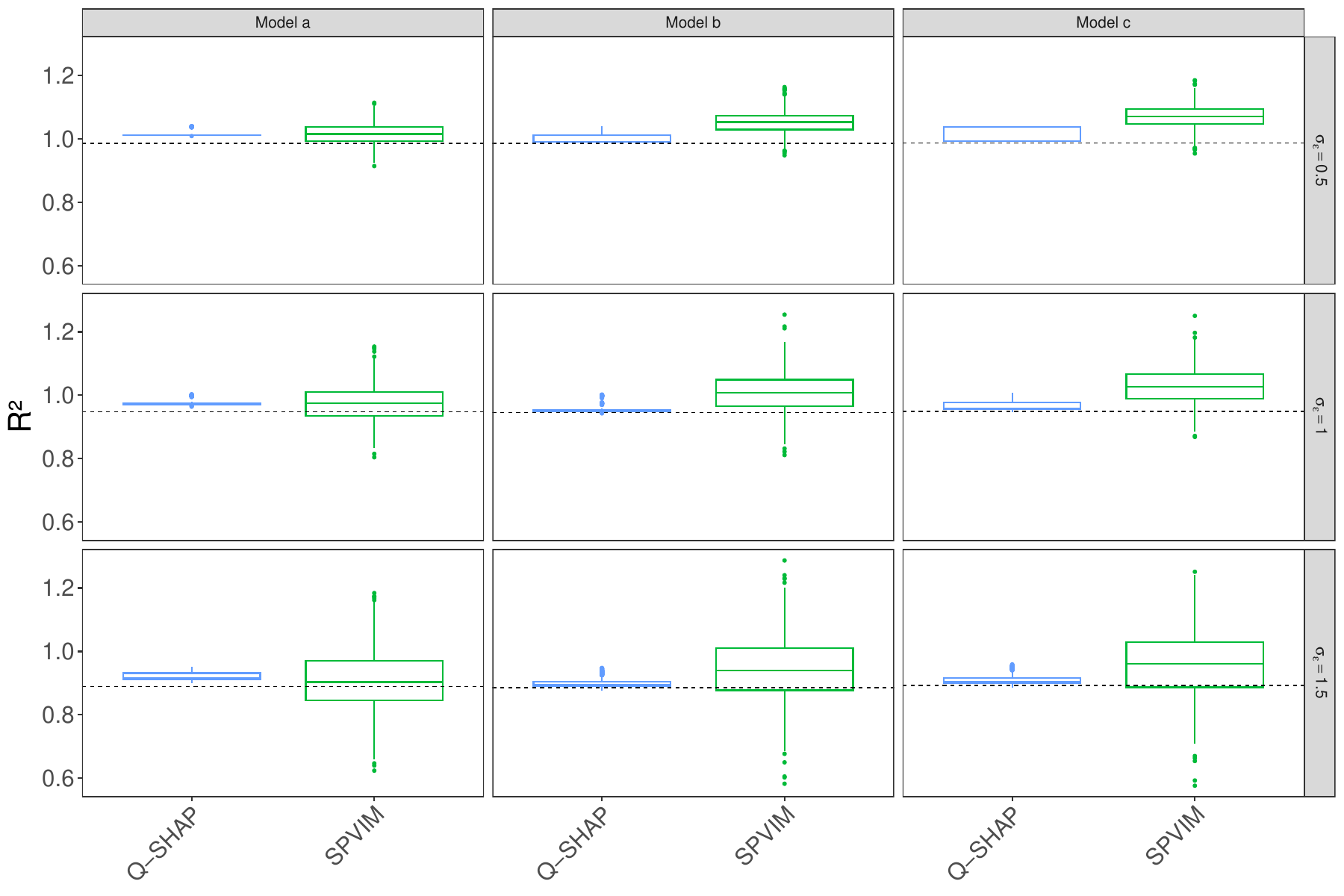}
    }
    \caption{Boxplots of (a) $X_1$-specific, (b) $X_2$-specific, (c) $X_3$-specific, and (d) the sum of all feature-specific $R^2$ in the three models with $n=2000$, $p=100$. The dashed lines show the theoretical $R^2$.}\label{np201}
\end{figure}

\begin{figure}[!hbp]
    \centering
    \subfigure[$X_1$-specific $R^2$]{
        \includegraphics[width=0.44\textwidth, height=4cm]{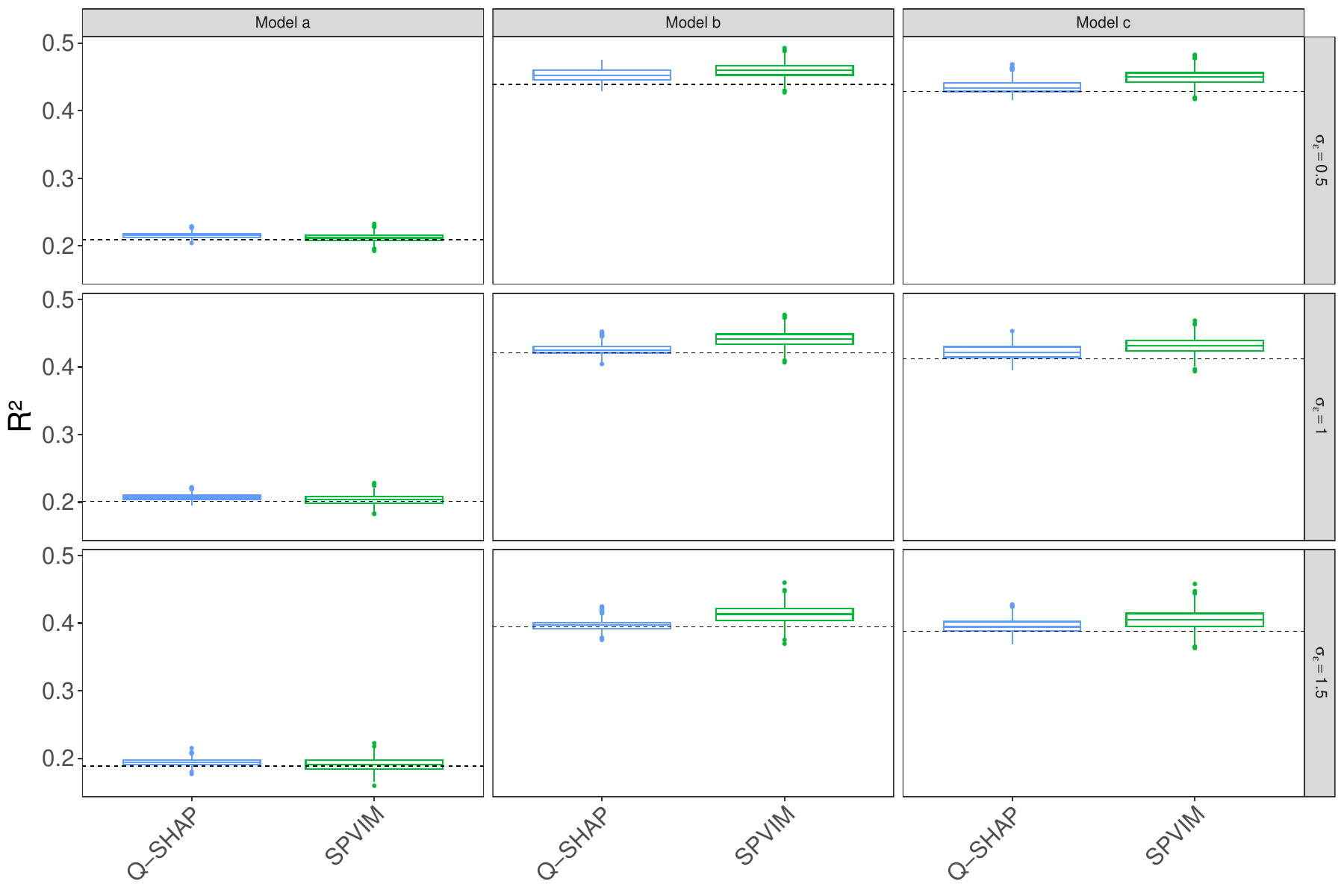}    }
    \subfigure[$X_2$-specific $R^2$]{
        \includegraphics[width=0.44\textwidth, height=4cm]{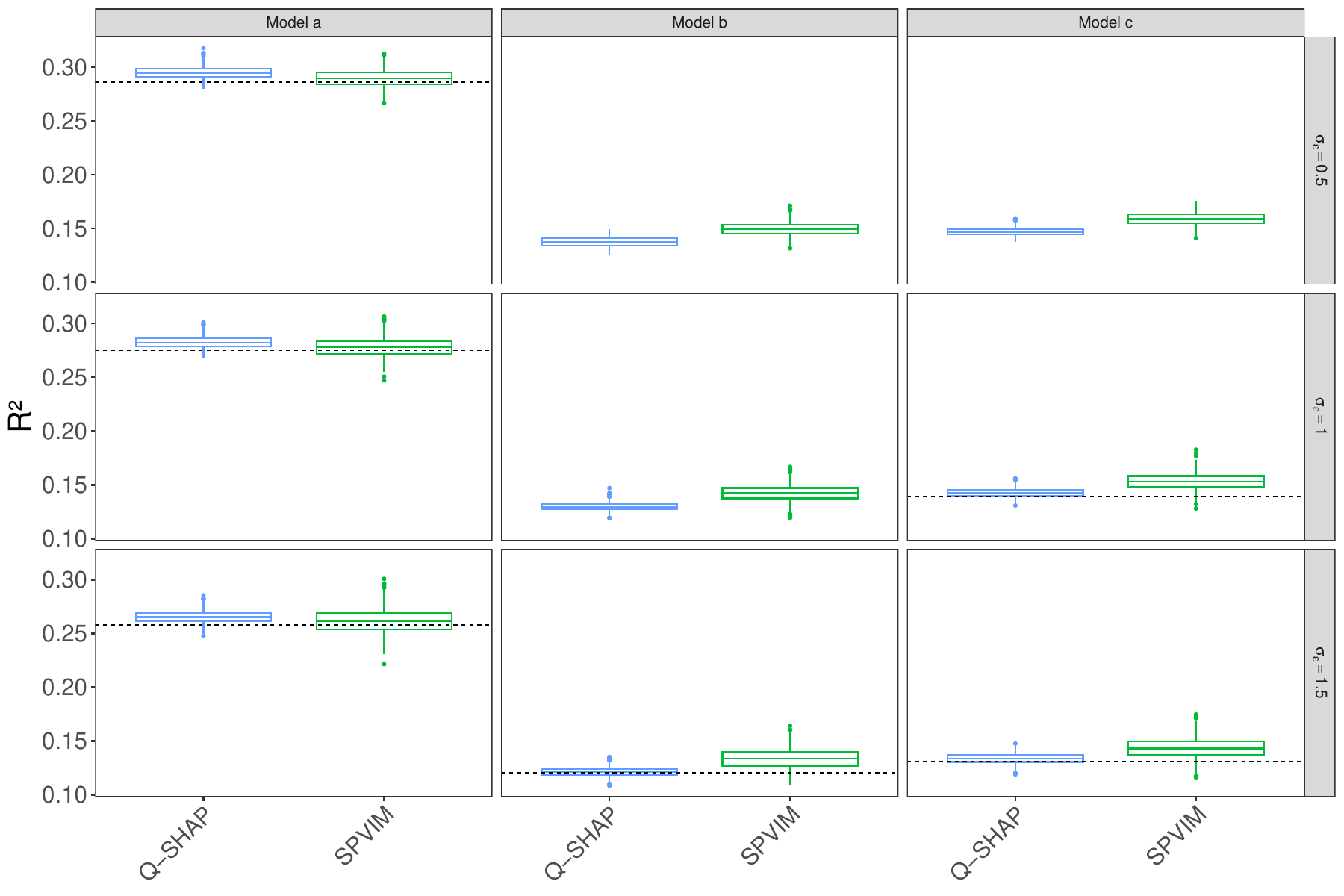}
    }
    
    \subfigure[$X_3$-specific $R^2$]{
        \includegraphics[width=0.44\textwidth, height=4cm]{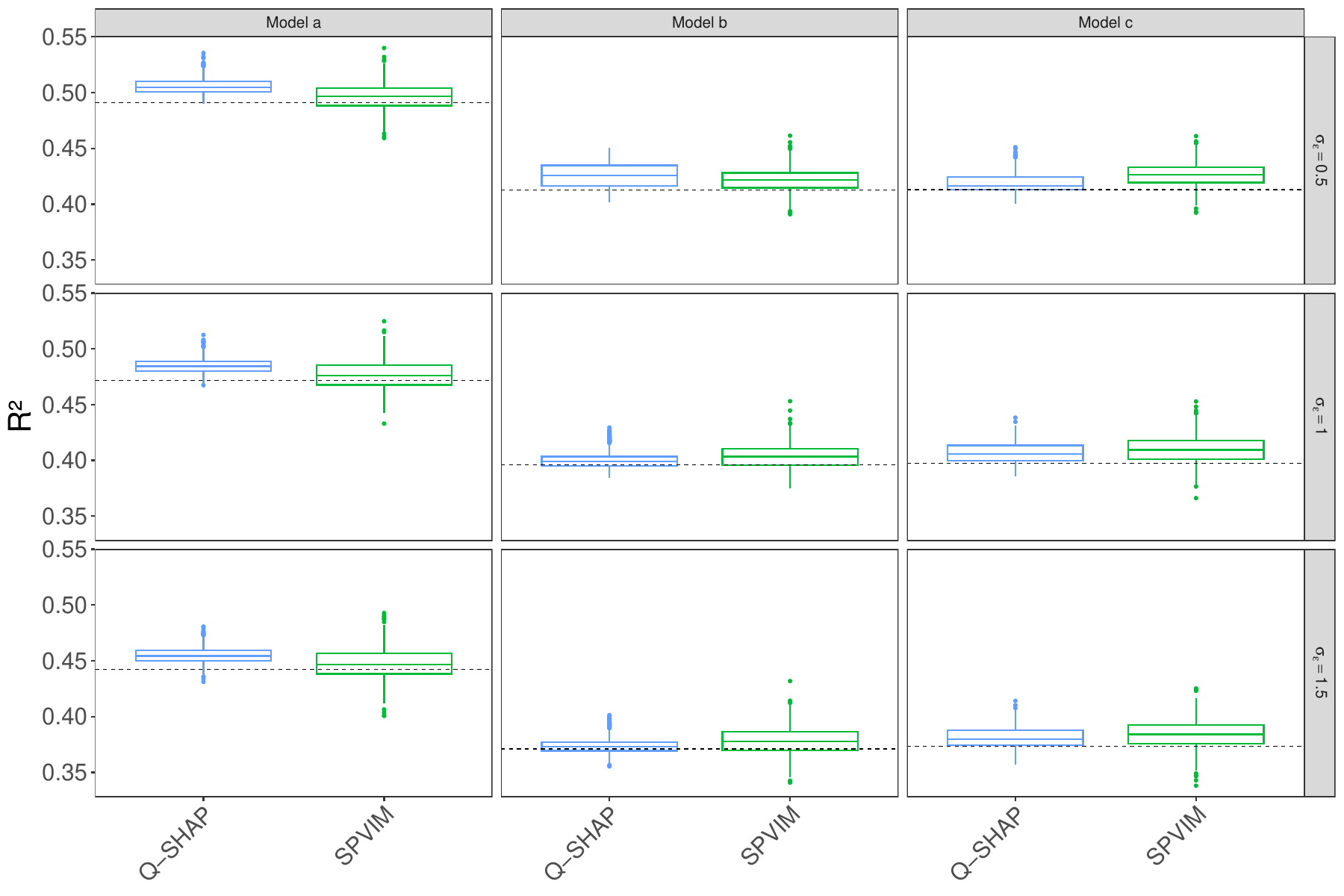}
    }
    \subfigure[Sum of all $R^2$]{
        \includegraphics[width=0.44\textwidth, height=4cm]{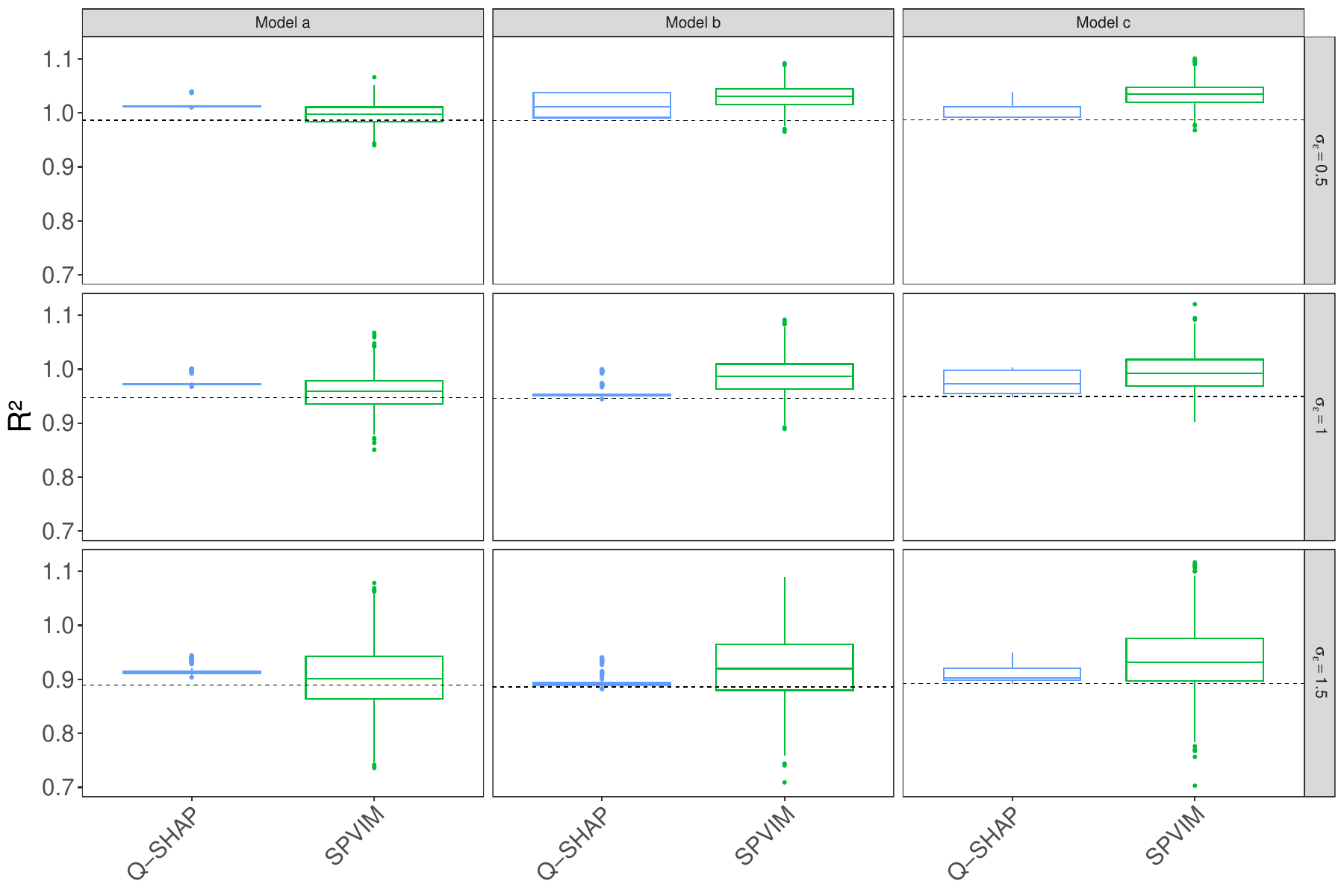}
    }
    \caption{Boxplots of (a) $X_1$-specific, (b) $X_2$-specific, (c) $X_3$-specific, and (d) the sum of all feature-specific $R^2$ in the three models with $n=5000$, $p=100$. The dashed lines show the theoretical $R^2$.}\label{np501}
\end{figure}

\begin{figure}[!htb]
    \centering
    \subfigure[$X_1$-specific $R^2$]{
        \includegraphics[width=0.44\textwidth, height=4cm]{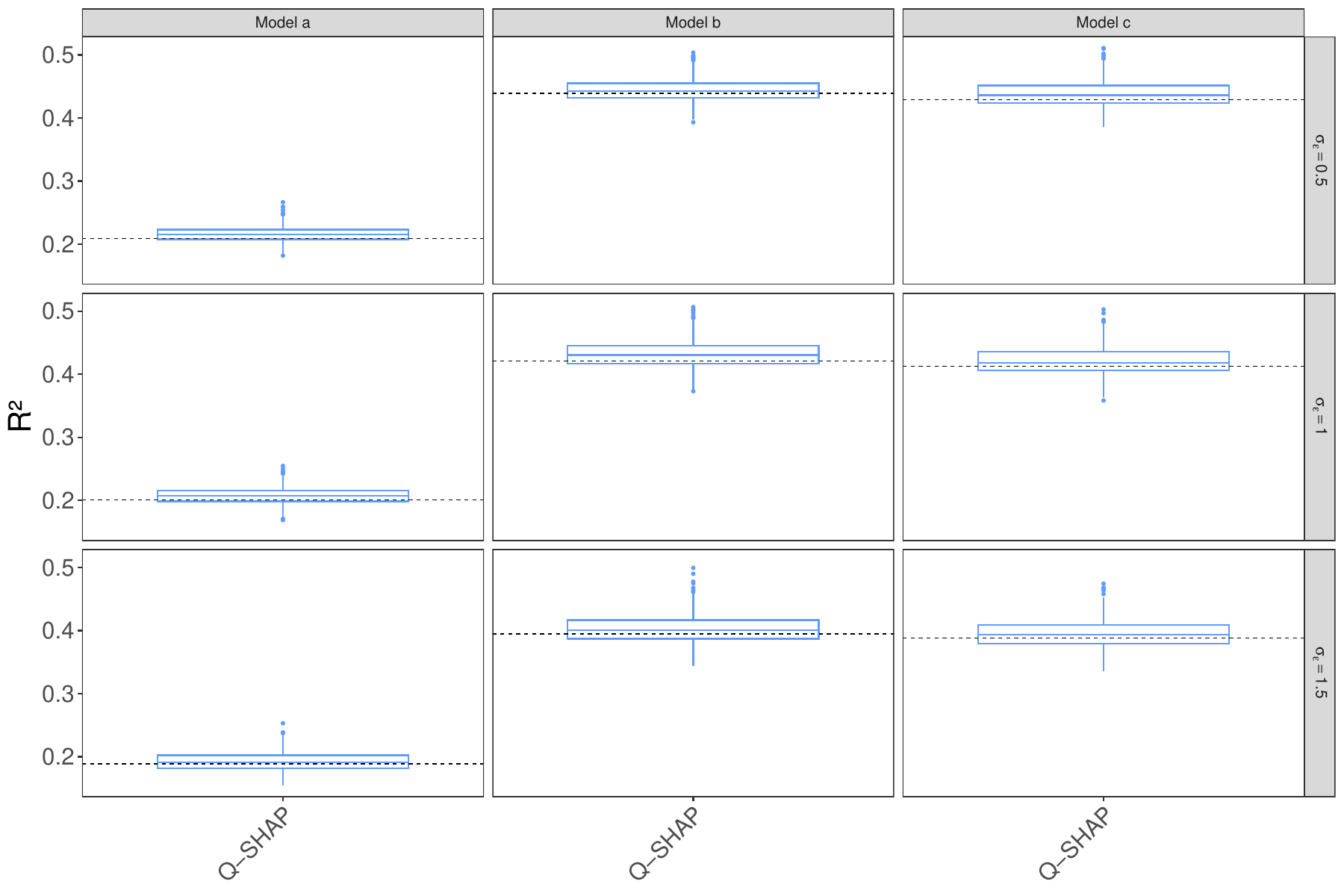}    }
    \subfigure[$X_2$-specific $R^2$]{
        \includegraphics[width=0.44\textwidth, height=4cm]{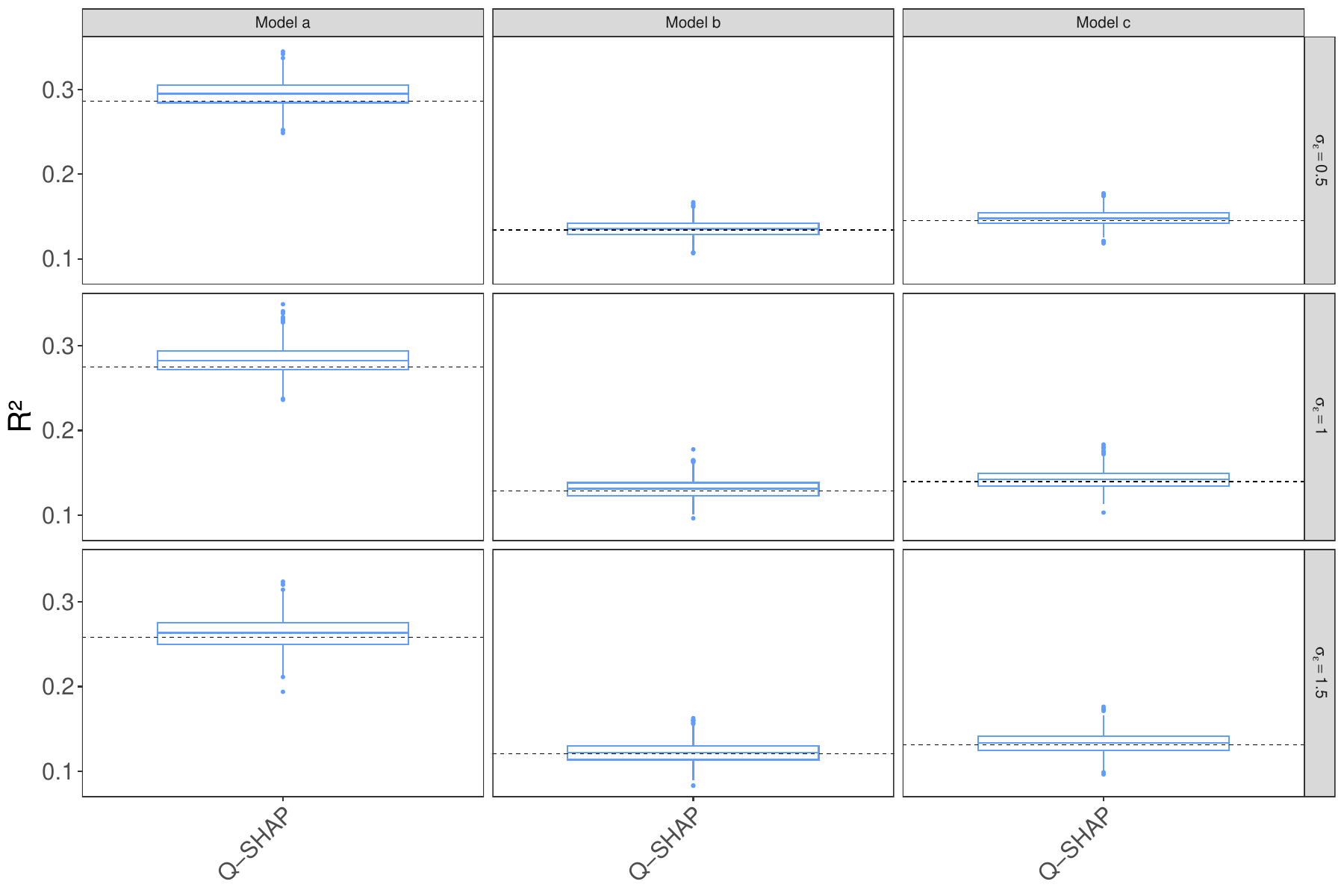}
    }
    
    \subfigure[$X_3$-specific $R^2$]{
        \includegraphics[width=0.44\textwidth, height=4cm]{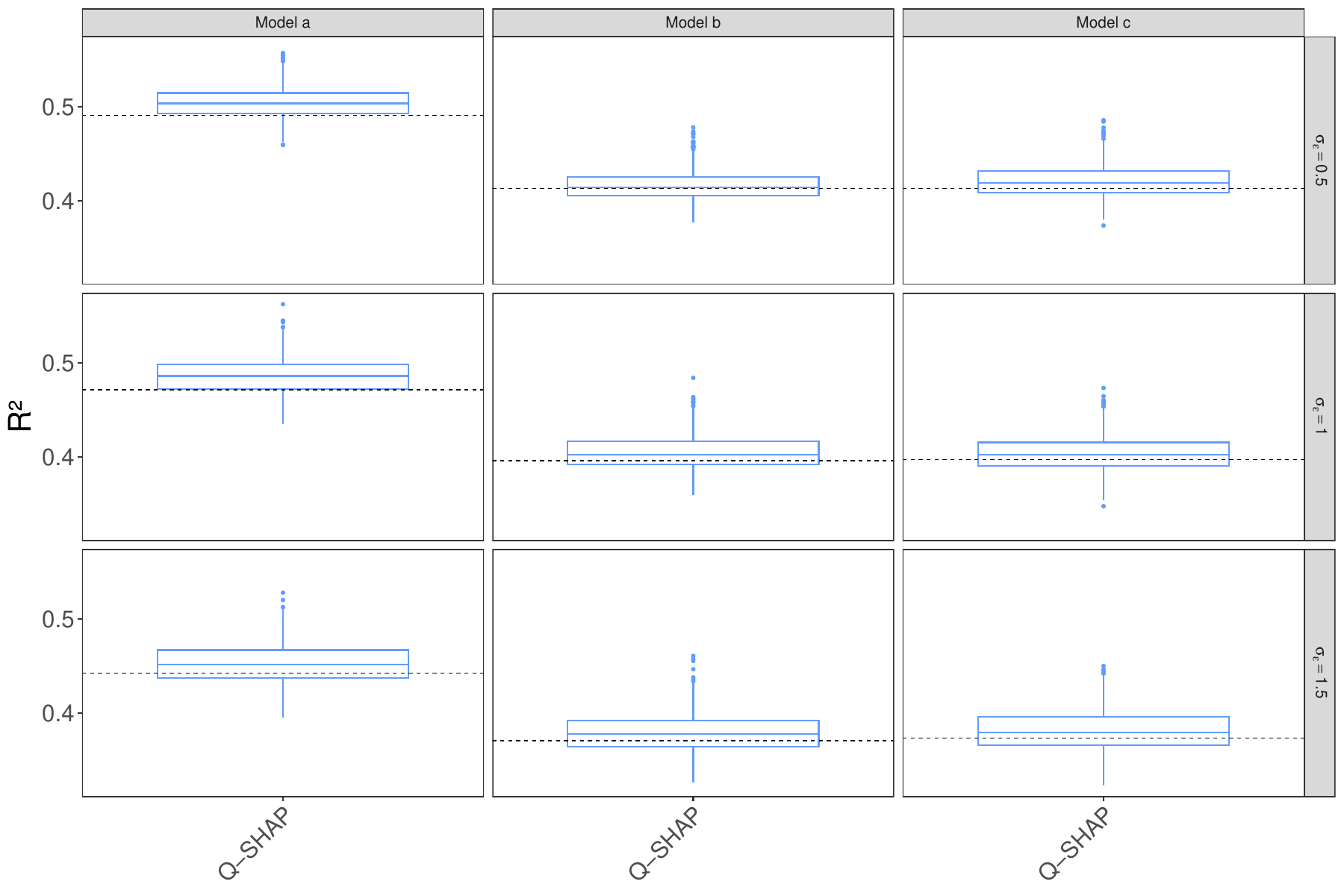}
    }
    \subfigure[Sum of all $R^2$]{
        \includegraphics[width=0.44\textwidth, height=4cm]{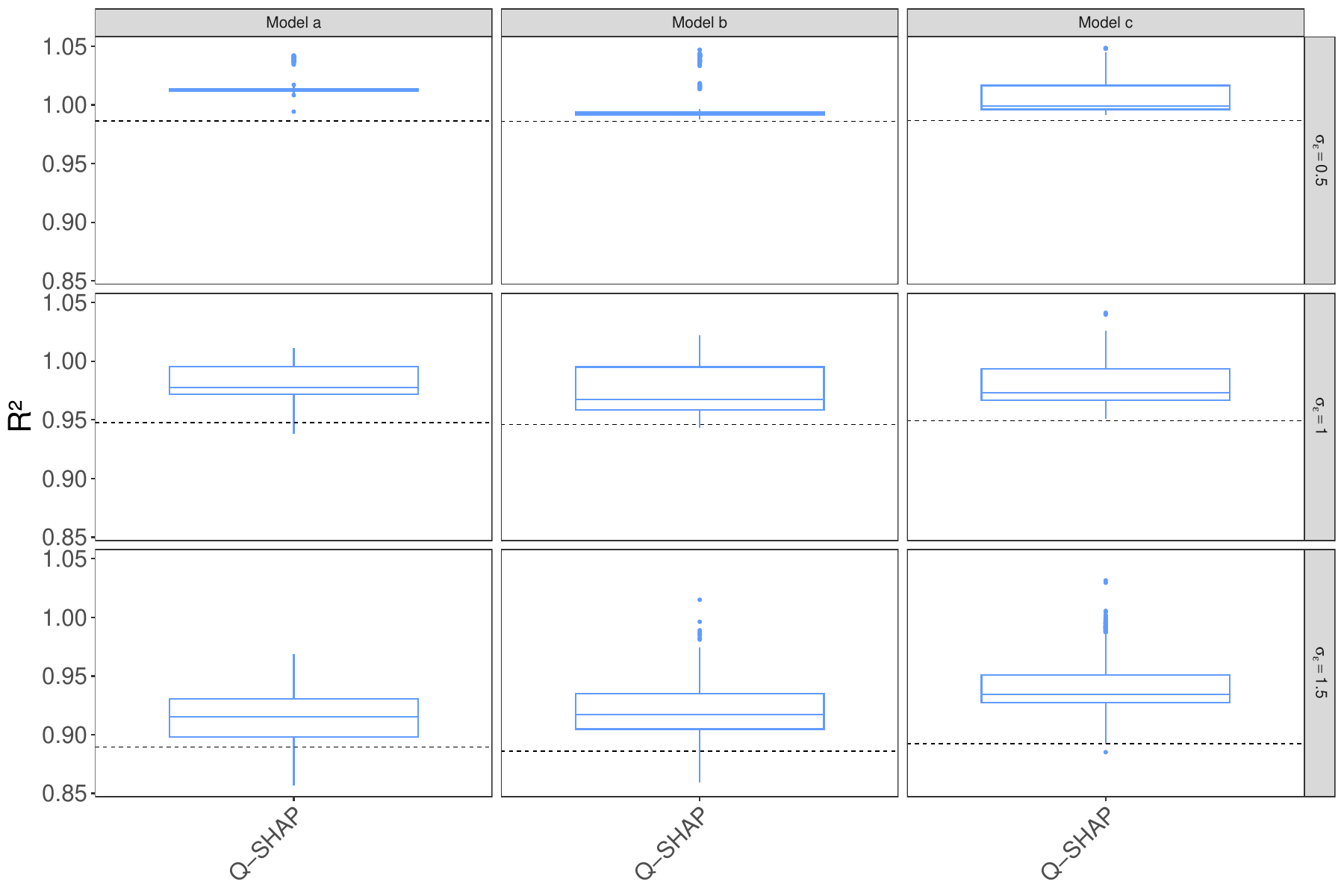}
    }
    \caption{Boxplots of (a) $X_1$-specific, (b) $X_2$-specific, (c) $X_3$-specific, and (d) the sum of all feature-specific $R^2$ in the three models with $n=500$, $p=500$. The dashed lines show the theoretical $R^2$.}\label{np055}
\end{figure}

\begin{figure}[!hbp]
    \centering
    \subfigure[$X_1$-specific $R^2$]{
        \includegraphics[width=0.44\textwidth, height=4cm]{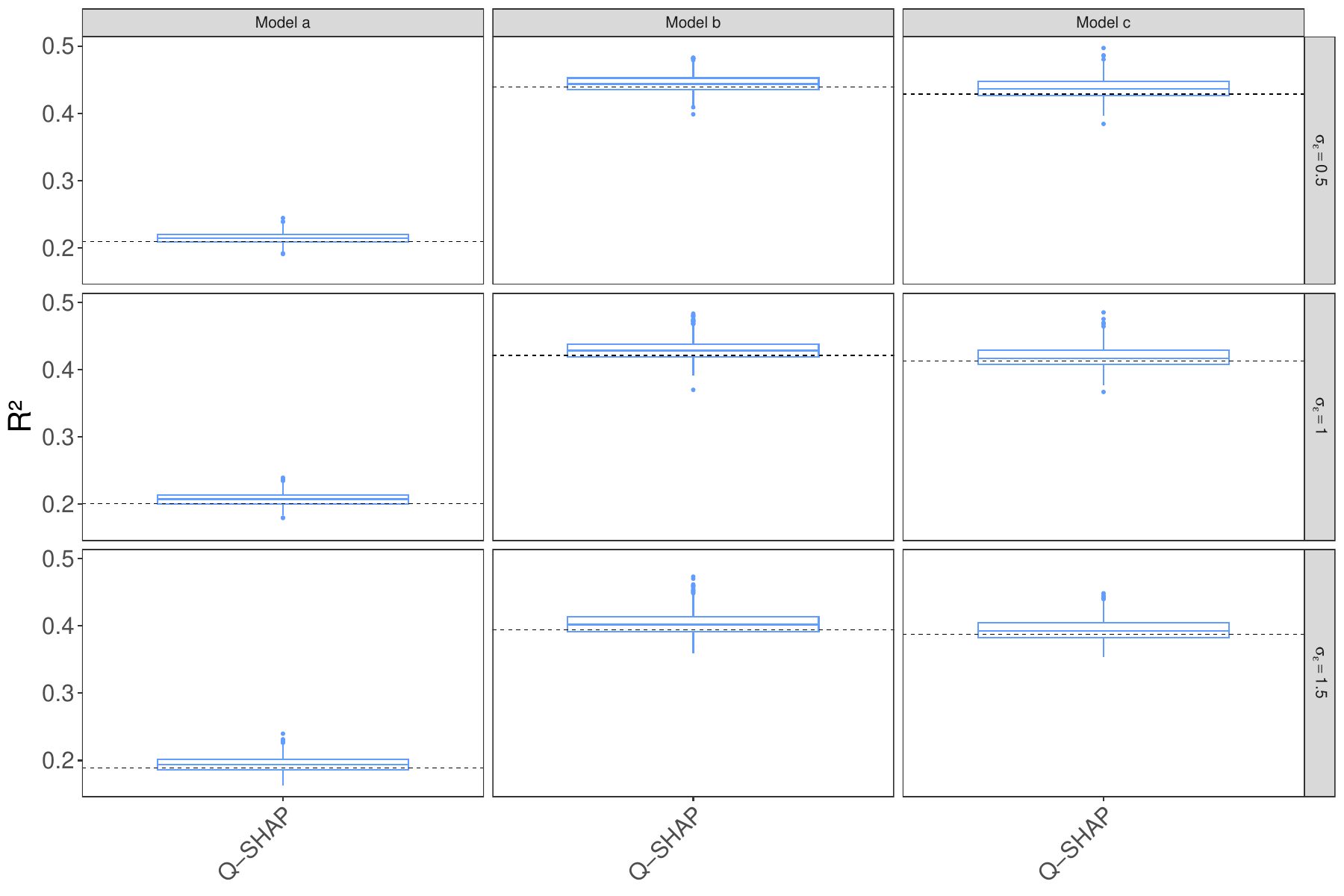}    }
    \subfigure[$X_2$-specific $R^2$]{
        \includegraphics[width=0.44\textwidth, height=4cm]{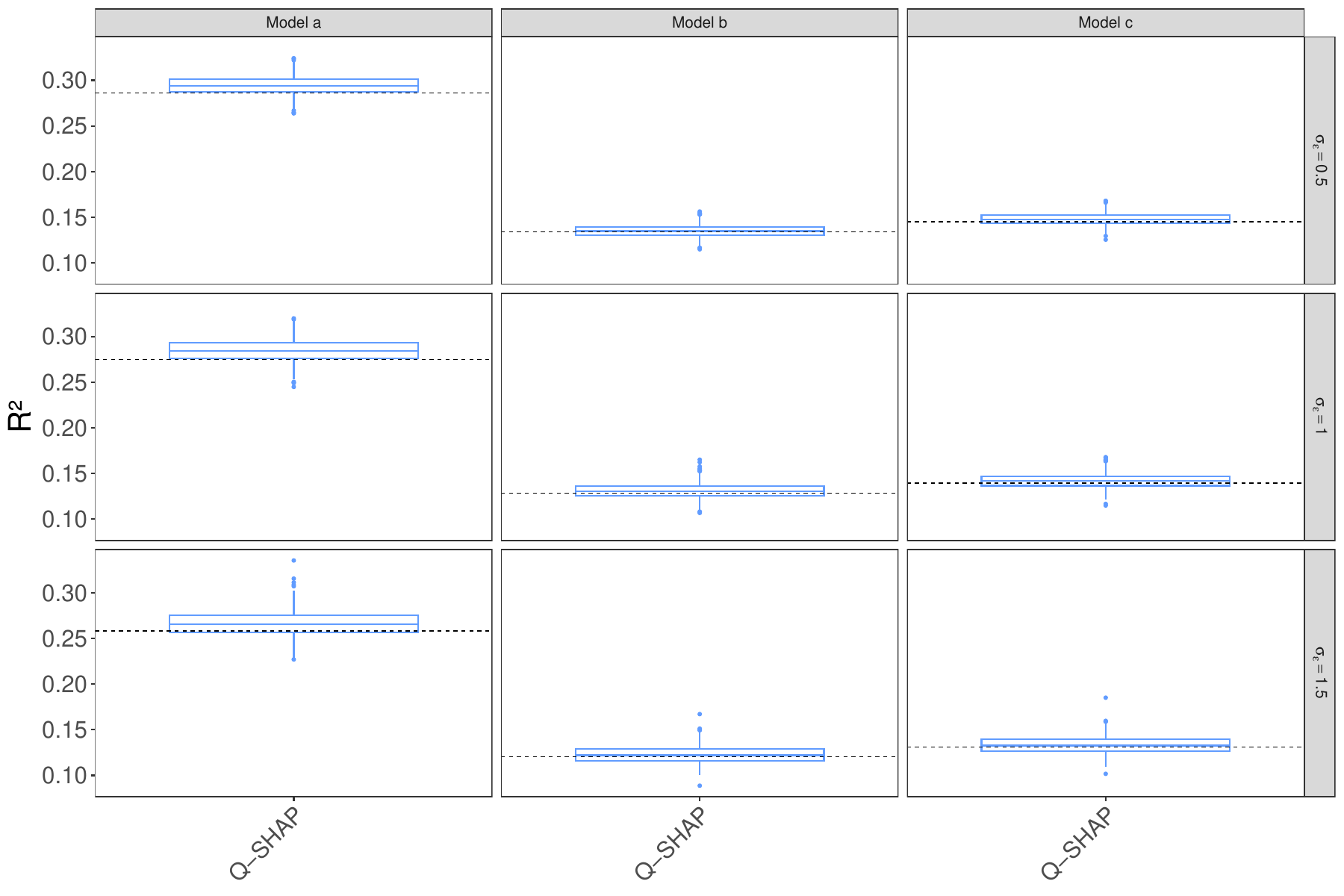}
    }
    
    \subfigure[$X_3$-specific $R^2$]{
        \includegraphics[width=0.44\textwidth, height=4cm]{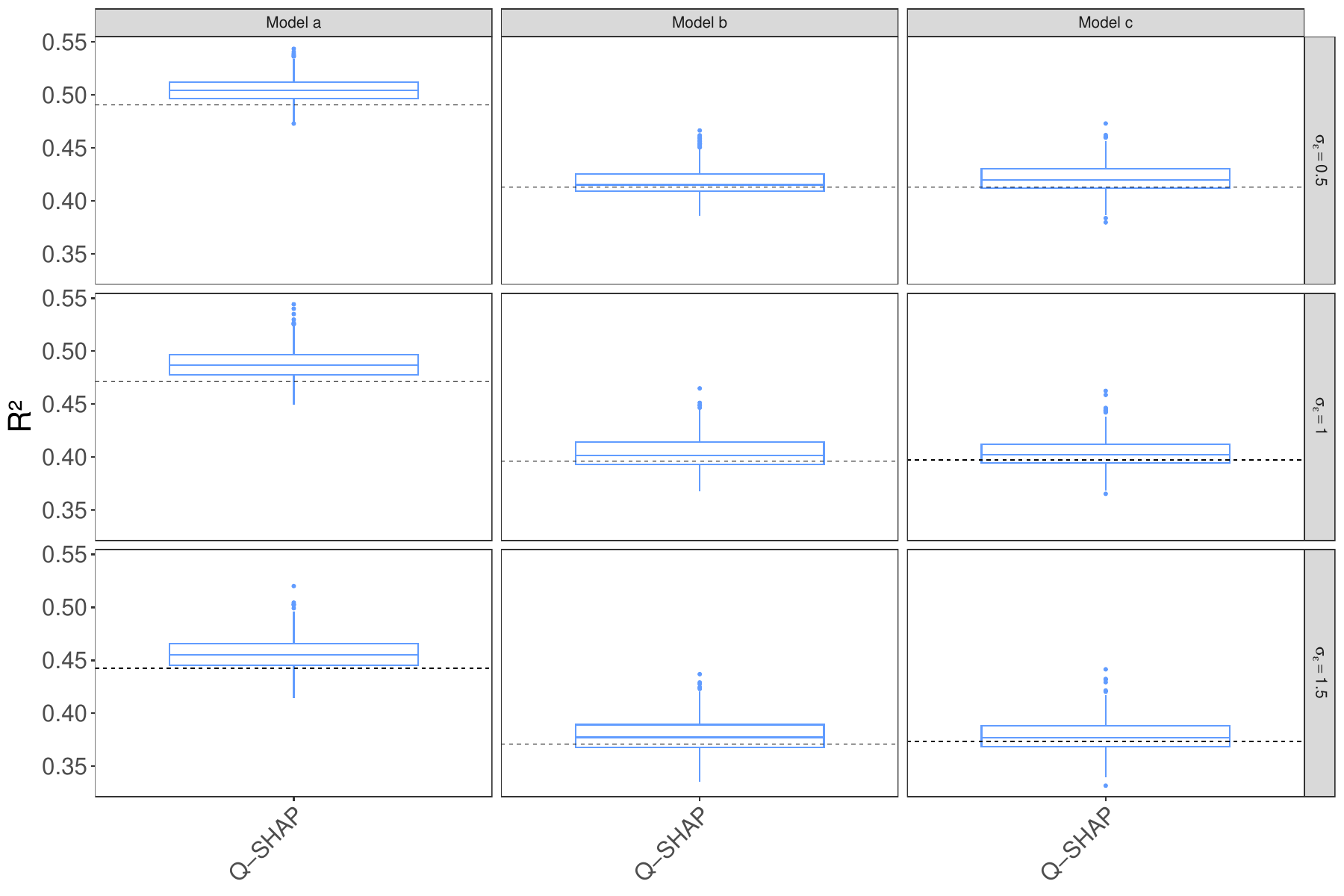}
    }
    \subfigure[Sum of all $R^2$]{
        \includegraphics[width=0.44\textwidth, height=4cm]{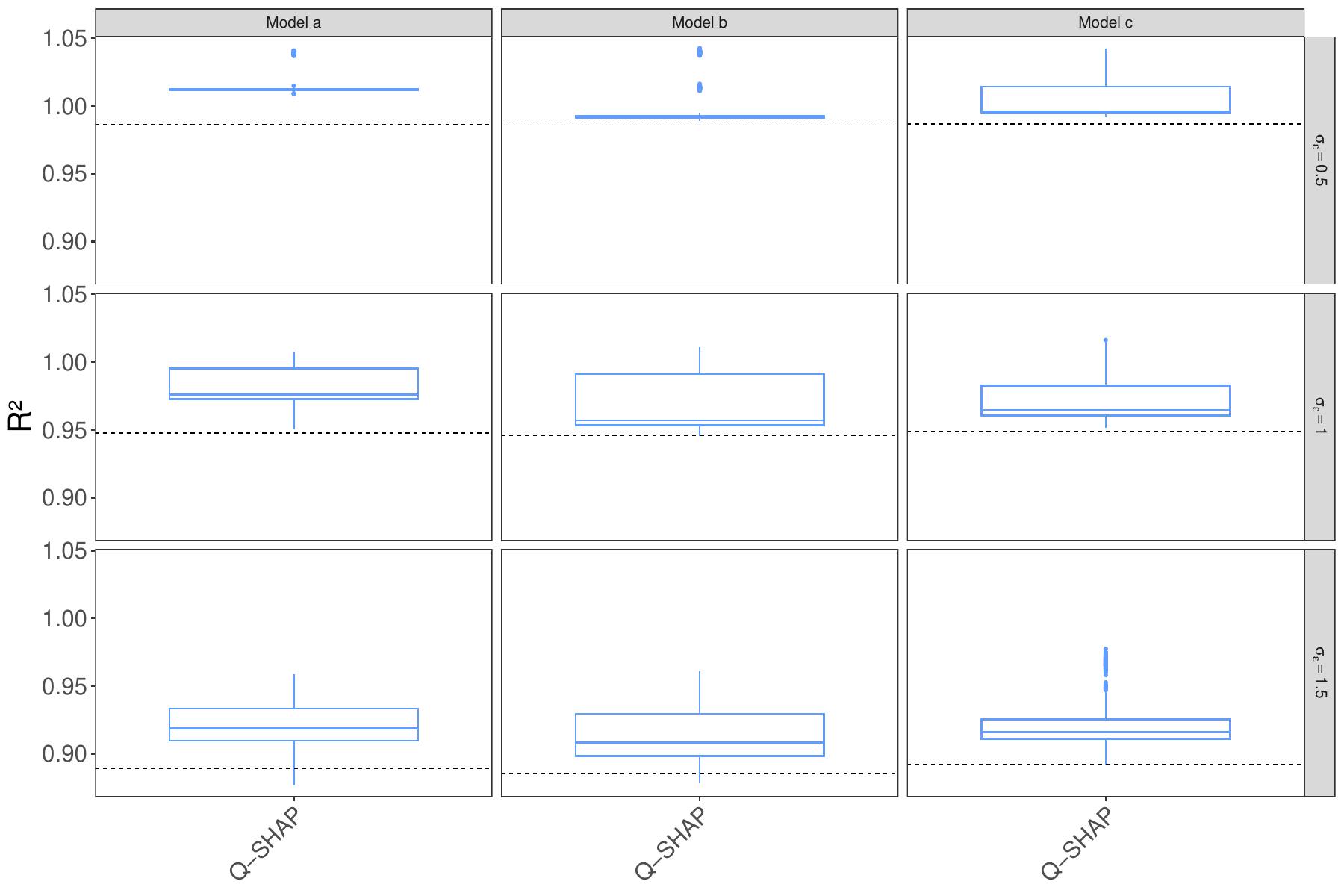}
    }
    \caption{Boxplots of (a) $X_1$-specific, (b) $X_2$-specific, (c) $X_3$-specific, and (d) the sum of all feature-specific $R^2$ in the three models with $n=1000$, $p=500$. The dashed lines show the theoretical $R^2$.}\label{np105}
\end{figure}

\begin{figure}[!htb]
    \centering
    \subfigure[$X_1$-specific $R^2$]{
        \includegraphics[width=0.44\textwidth, height=4cm]{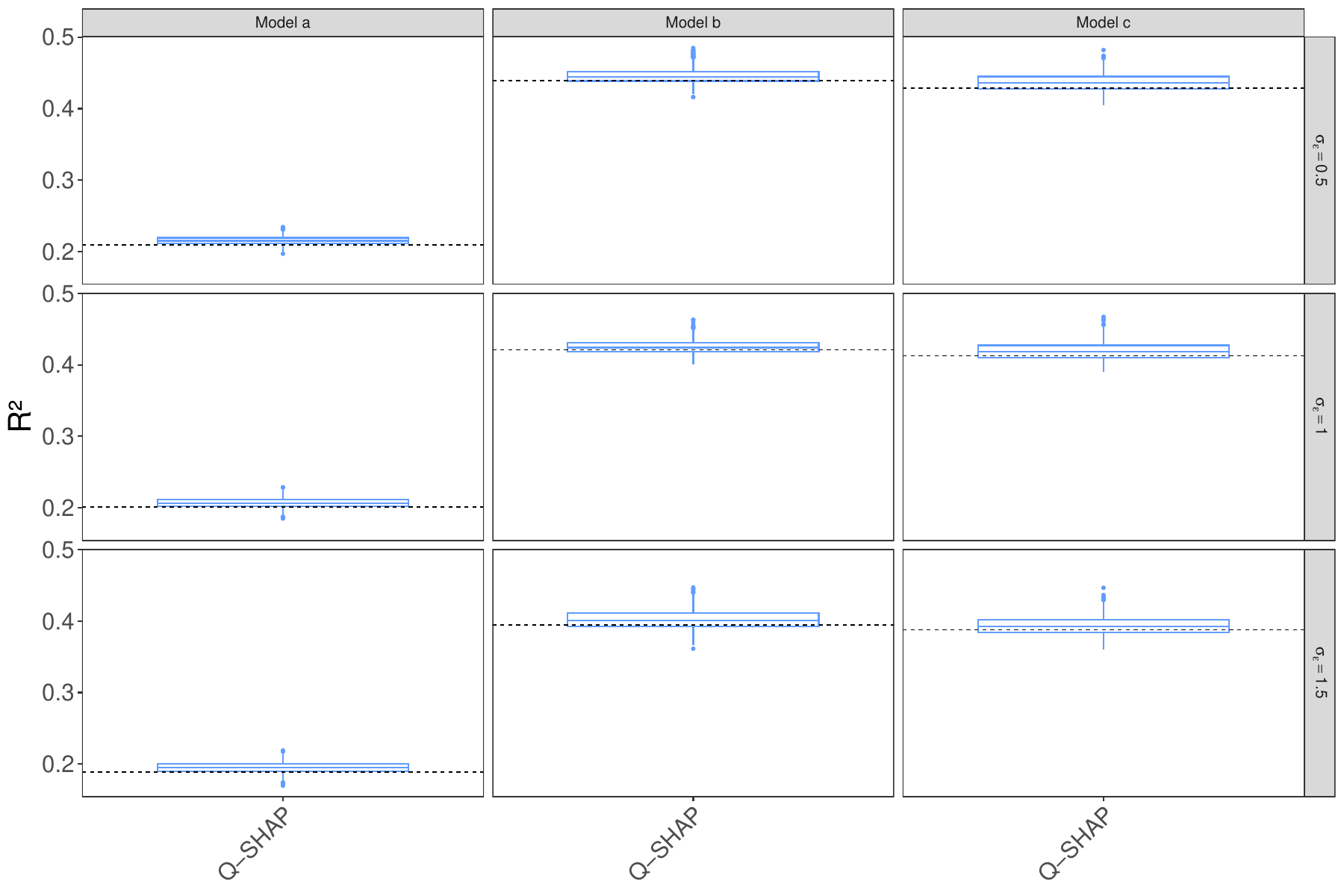}    }
    \subfigure[$X_2$-specific $R^2$]{
        \includegraphics[width=0.44\textwidth, height=4cm]{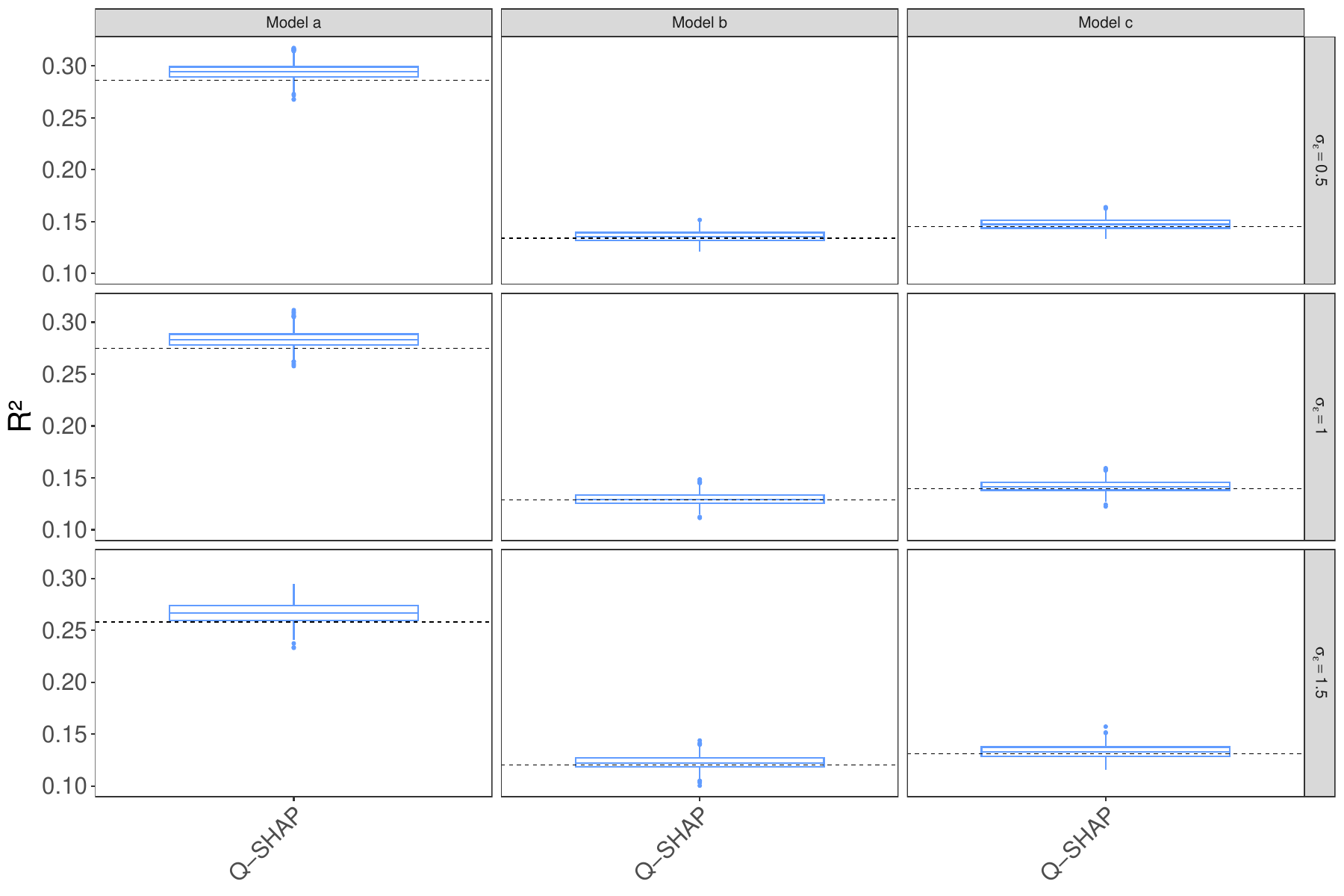}
    }
    
    \subfigure[$X_3$-specific $R^2$]{
        \includegraphics[width=0.44\textwidth, height=4cm]{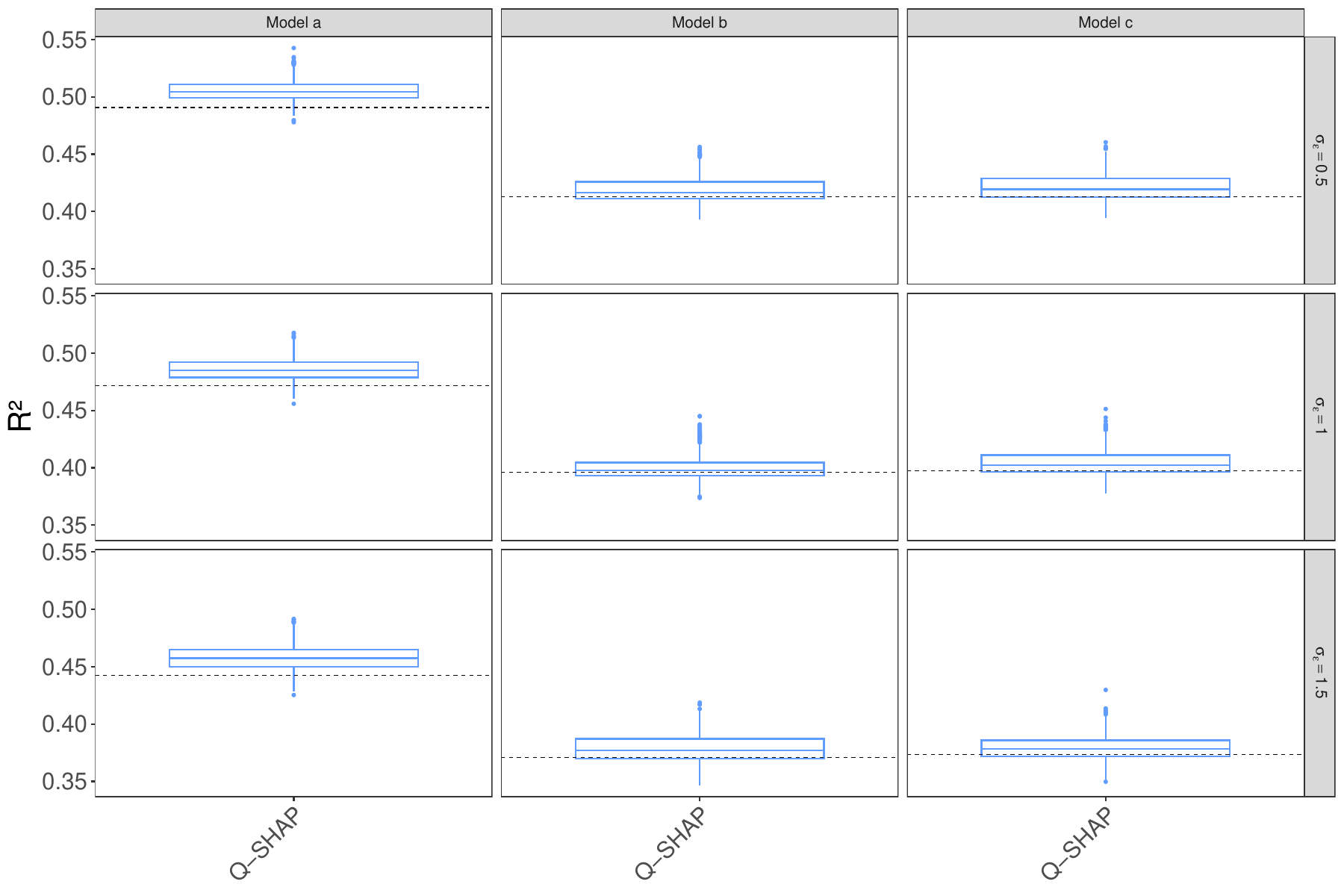}
    }
    \subfigure[Sum of all $R^2$]{
        \includegraphics[width=0.44\textwidth, height=4cm]{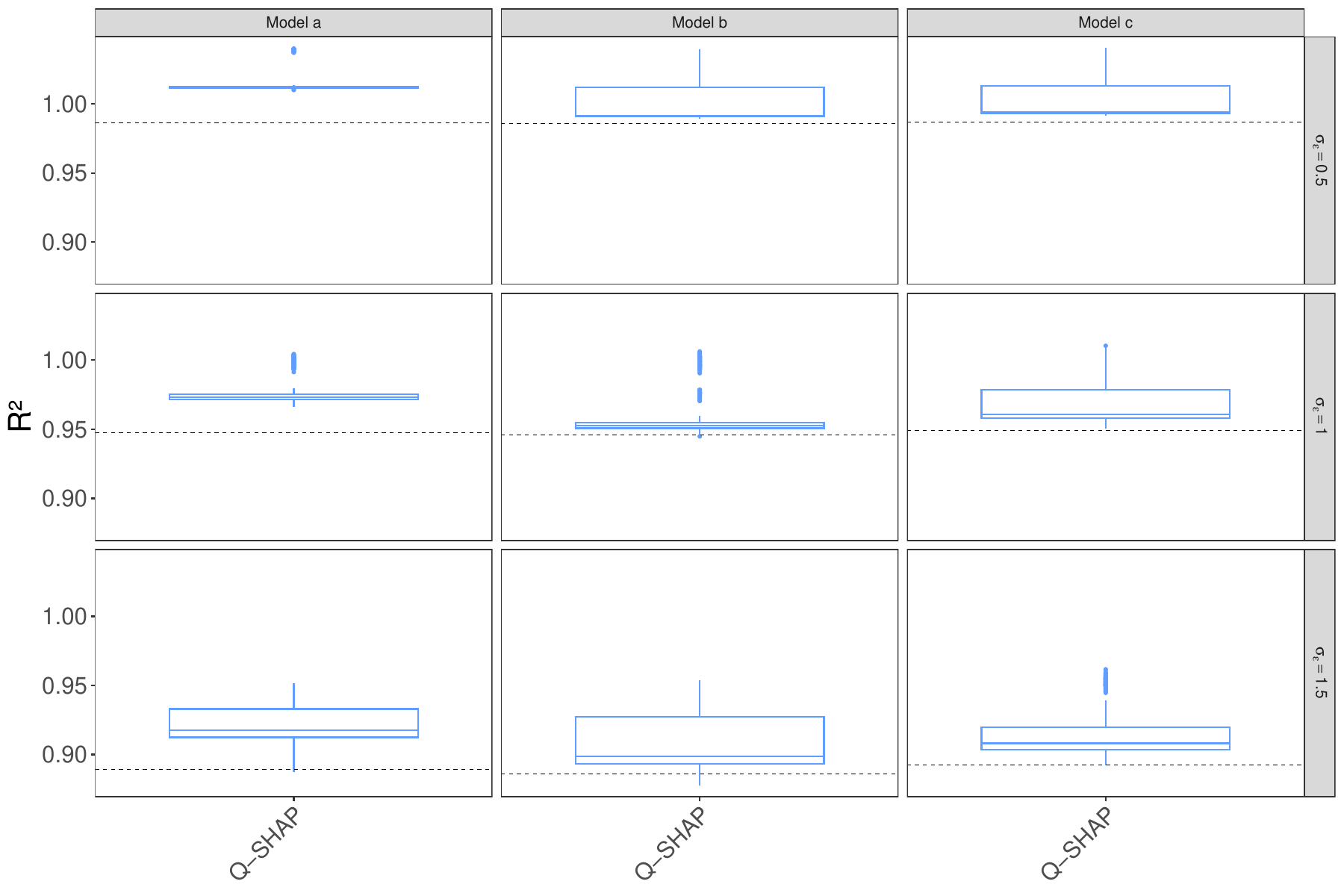}
    }
    \caption{Boxplots of (a) $X_1$-specific, (b) $X_2$-specific, (c) $X_3$-specific, and (d) the sum of all feature-specific $R^2$ in the three models with $n=2000$, $p=500$. The dashed lines show the theoretical $R^2$.}\label{np205}
\end{figure}

\begin{figure}[!hbp]
    \centering
    \subfigure[$X_1$-specific $R^2$]{
        \includegraphics[width=0.44\textwidth, height=4cm]{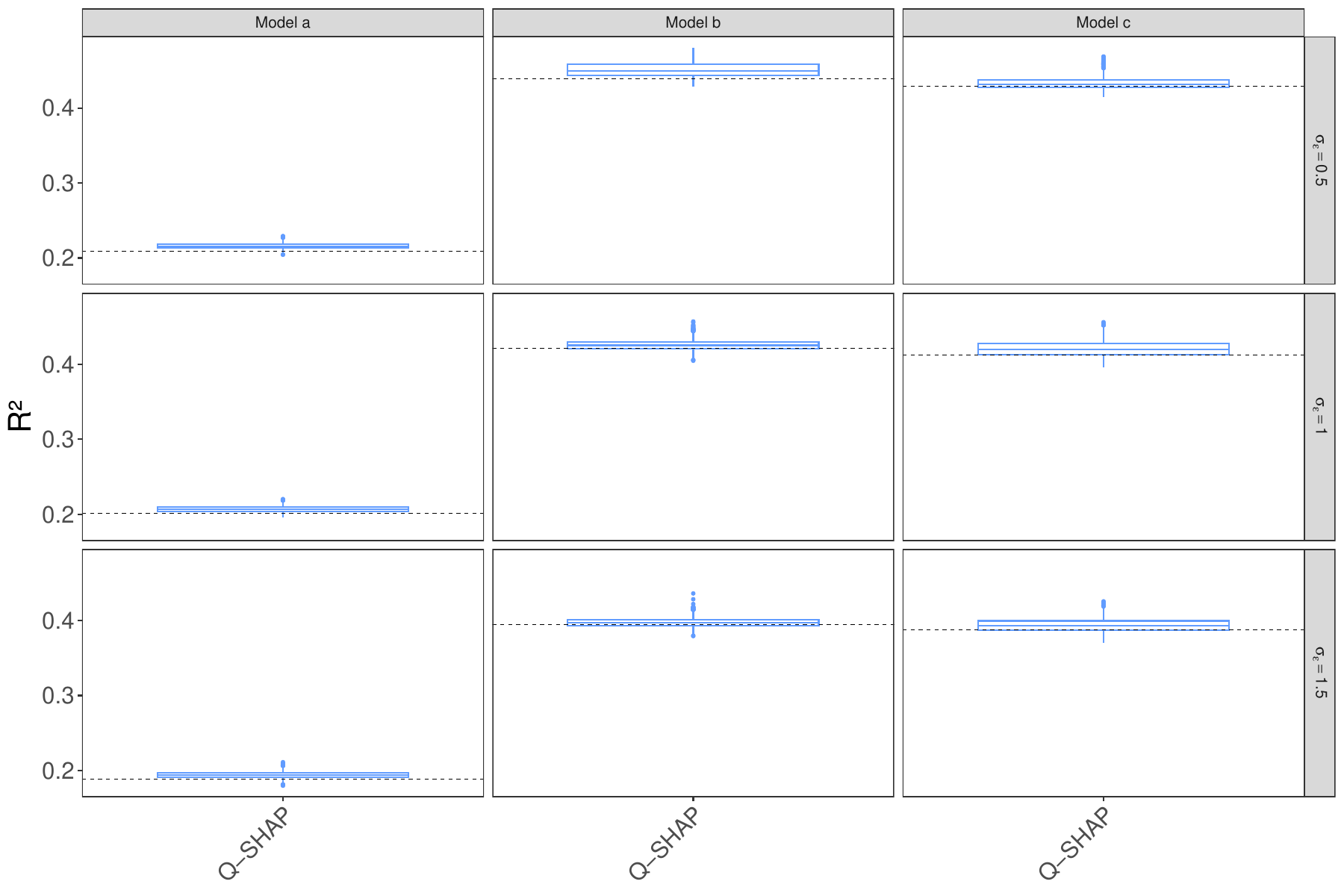}    }
    \subfigure[$X_2$-specific $R^2$]{
        \includegraphics[width=0.44\textwidth, height=4cm]{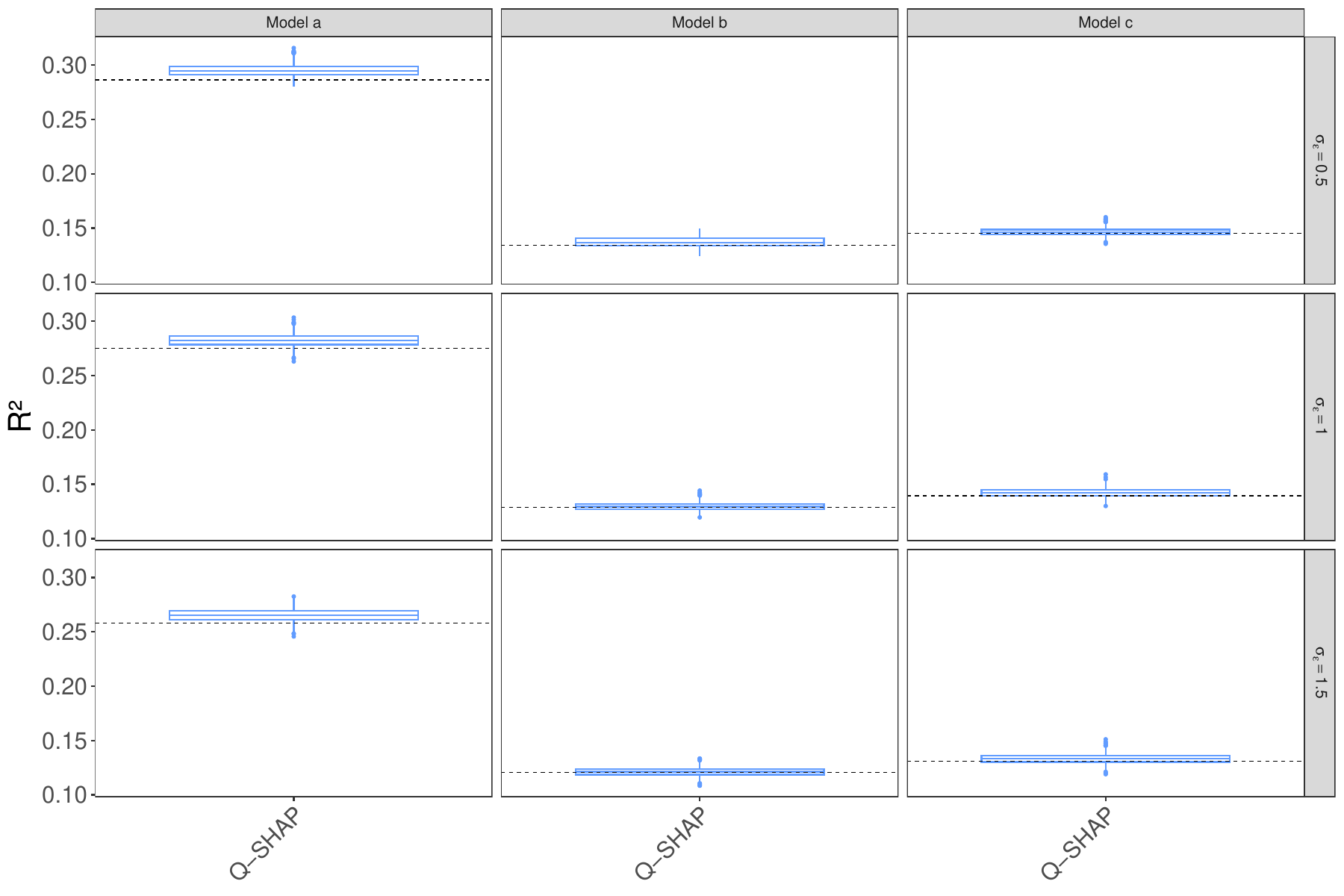}
    }
    
    \subfigure[$X_3$-specific $R^2$]{
        \includegraphics[width=0.44\textwidth, height=4cm]{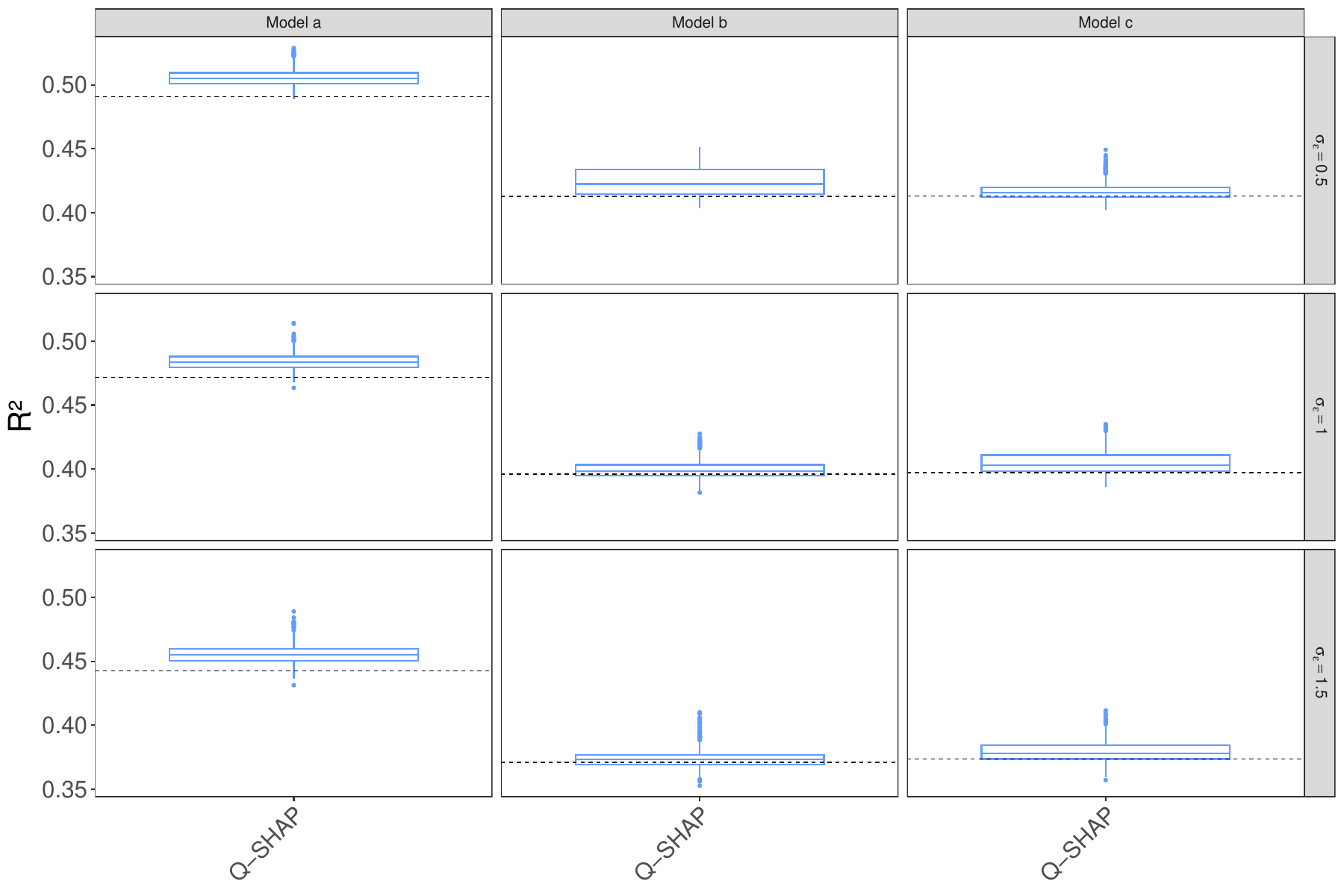}
    }
    \subfigure[Sum of all $R^2$]{
        \includegraphics[width=0.44\textwidth, height=4cm]{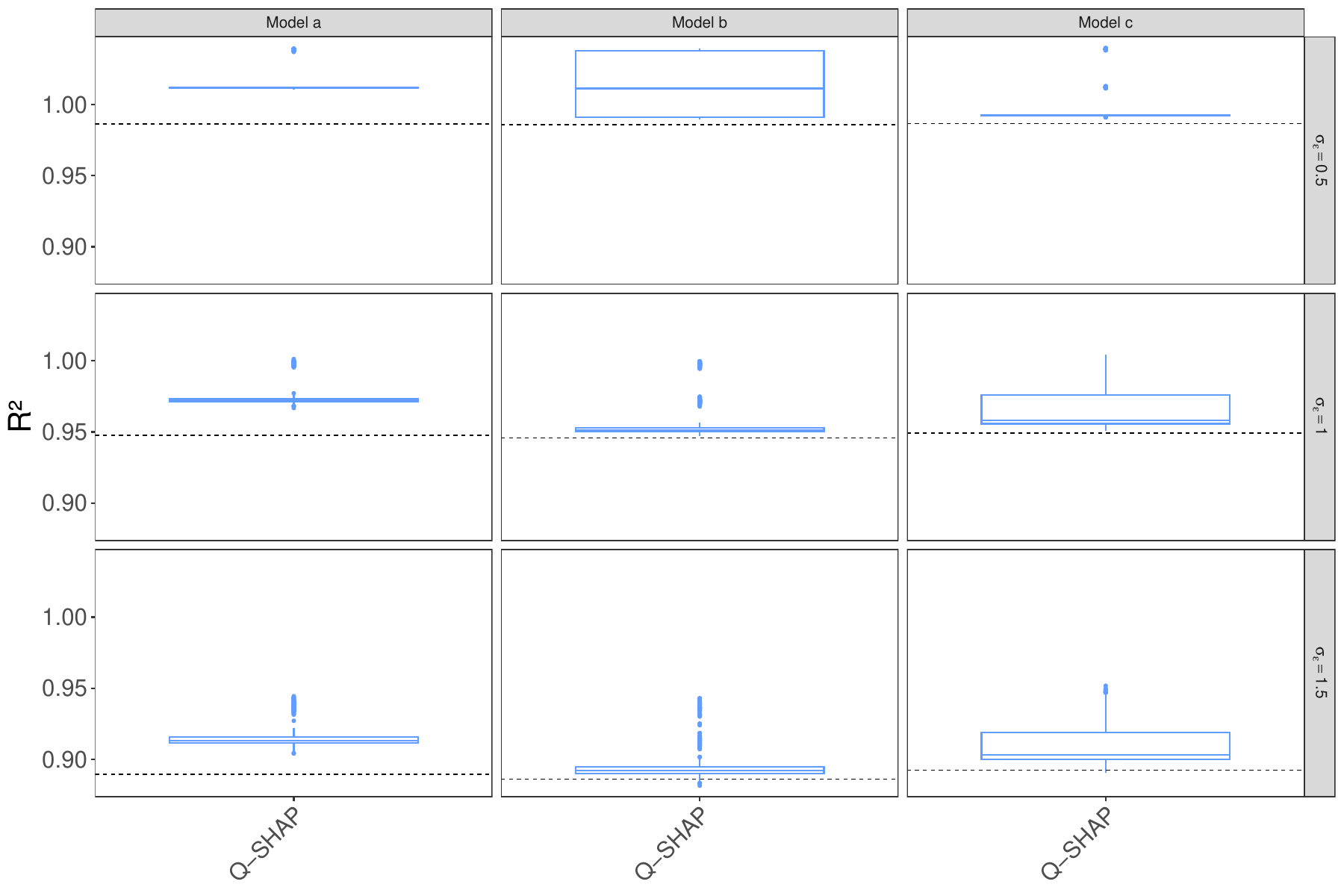}
    }
    \caption{Boxplots of (a) $X_1$-specific, (b) $X_2$-specific, (c) $X_3$-specific, and (d) the sum of all feature-specific $R^2$ in the three models with $n=5000$, $p=500$. The dashed lines show the theoretical $R^2$.}\label{np505}
\end{figure}

We have compared the performance of three different methods, i.e., our proposed Q-SHAP, SAGE by \citet{covert2020understanding}, and SPVIM by \citet{williamson2020efficient}, in calculating the feature-specific $R^2$ as well as the sum of all feature-specific $R^2$ for the three models specified in Section~\ref{simulation_study}, with different settings, i.e., $n\in\{500, 1000, 2000, 5000\}$, $p\in\{100,500\}$, and $\sigma_{\epsilon}\in\{0.5, 1, 1.5\}$. The results are shown in Fig.~\ref{np051}-\ref{np505}. Note that the results of SAGE are unavailable in Fig.~\ref{np201}-\ref{np505} because it cannot report those $R^2$ with our limited computational resources, and the results of SPVIM are unavailable in Fig.~\ref{np055}-\ref{np505} because it demands too much time to complete the computation when $p=500$. 

\subsection{Plots of the mean absolute error (MAE)} \label{SA_Simu_MAE}

\begin{figure}[!htb]
    \centering 
    \includegraphics[scale=0.48]{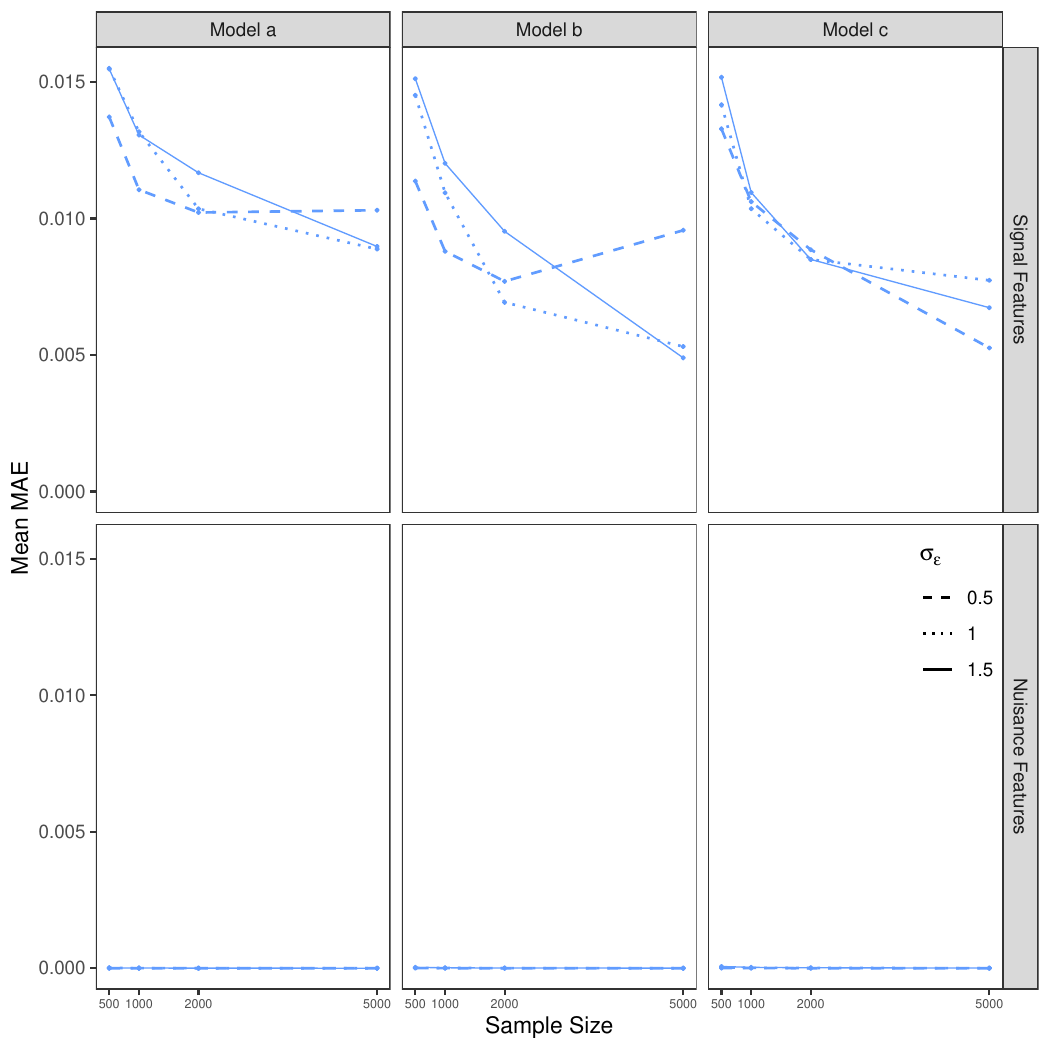}
    \caption{The mean of absolute error (MAE) of the feature-specific $R^2$ by Q-SHAP averaged across 1,000 datasets with $p=500$}\label{mmae_500}
\end{figure}

Similar to Fig.~\ref{mmae}, we show in Fig.~\ref{mmae_500} the mean absolute error (MAE) of feature-specific $R^2$ for both signal and nuisance features averaged over 1,000 datasets when $p=500$. Note that the results of SAGE and SPVIM are unavailable because none of them can complete the computation for $p=500$ with limited computational resources.

\section{Complexity of the Algorithm}\label{complexity_algo}

Here we assume that the dataset includes $n$ samples as well as $p$ features, and a total of $T$ trees are constructed with the maximum tree depth $D$ and maximum tree leaves $L$. We denote $S$ the number of permutations taken in SAGE with $S=10^{20}$ by default. Then, when the trees are constructed by XGBoost, the complexity of SAGE is $O(TDSpn)$ \citep{covert2020understanding}, and the complexity of SPVIM is $O(TDpn^2 \log n)$ \citep{williamson2020efficient}. Instead, the complexity of Q-SHAP is $O(T L^2 D^2 n)$ which doesn't rely on the number of features $p$. 

Let's first consider the complexity of Q-SHAP in  Algorithm 1 for a single tree and one sample. As shown in Algorithm~\ref{q-shap}, the two outer loops that iterate through the tree leaves, result in a complexity of $O(L^2)$. Within the inner loop, the computation involves the number of features induced by each pair of leaves, leading to $O(D)$ operations. The evaluation of $t[j]$  involves the computation of  $C(z) \cdot P(z)$, which takes $O(D)$ operations since the number of union features between two leaves is bounded by $2D$. Combining these, the overall complexity for one tree and one sample is $O(L^2D^2)$. Thus, for the whole dataset, the complexity of Q-SHAP scales to $O(nL^2D^2)$ for a single tree. With the advancements introduced in Section 4, Q-SHAP has a total complexity of $O(TnL^2D^2)$ for the ensemble of $T$ boosting trees.

The property that the complexity of Q-SHAP doesn't rely on the number of features is a prominent advantage of Q-SHAP and is critical in analyzing high-dimensional data. Such an advantage is achieved via Proposition \ref{dimension_reduce}, which eliminates dependence on $p$ by leveraging the internal structure of the tree. Furthermore, unlike SAGE and SPVIM, which require extensive sampling, Q-SHAP directly utilizes the tree’s weight function introduced in Section 3.2, eliminating the need for any sampling.

\end{document}